\pdfoutput=1
\documentclass[11pt]{article}

\usepackage[final]{acl}

\usepackage{times}
\usepackage{latexsym}

\usepackage[T1]{fontenc}
\usepackage[utf8]{inputenc}

\usepackage{microtype}

\usepackage{inconsolata}

\usepackage{graphicx}

\usepackage{amsmath}
\usepackage{amsthm}
\usepackage{amsfonts}
\usepackage{enumitem}
\usepackage{xspace}

\newcommand\cites[1]{\citeauthor{#1}'s\ (\citeyear{#1})}

\newcommand\refhyp[0]{\hyperref[hyp1]{LKCH}\xspace}

\newtheoremstyle{named}{}{}{\itshape}{}{\bfseries}{:}{.5em}{\thmnote{#3 }#1 (LKCH)}
\theoremstyle{named}
\newtheorem{hypothesis}{Hypothesis}

\title{Exploring Graph Representations of Logical Forms for Language Modeling}

\author{Michael Sullivan \\
  University at Buffalo \\
  Saarland University \\
  \texttt{msullivan@lst.uni-saarland.de}}

\begin{document}
\maketitle
\begin{abstract}
We make the case for language models over logical forms (LFLMs), arguing that such models are more data-efficient than their textual counterparts. To that end, we introduce the \textit{\underline{G}raph-based \underline{Fo}rmal-\underline{L}ogical \underline{D}istributional \underline{S}emantics} (GFoLDS) prototype, a pretrained LM over graph representations of logical forms, as a proof-of-concept of LFLMs. Using GFoLDS, we present strong experimental evidence that LFLMs can leverage the built-in, basic linguistic knowledge inherent in such models to immediately begin learning more complex patterns. On downstream tasks, we show that GFoLDS vastly outperforms textual, transformer LMs (BERT) pretrained on the same data, indicating that LFLMs can learn with substantially less data than models over plain text. Furthermore, we show that the performance of this model is likely to scale with additional parameters and pretraining data, suggesting the viability of LFLMs in real-world applications.
\end{abstract}

\section{Introduction}
\label{sec_intro}

Although recent advances in LLMs have led to remarkable performance on a wide variety of benchmarks, the consistent improvements exhibited by SoTA LLMs are largely due to corresponding increases in model size \cite{villalobos2022will,muennighoff2024scaling}. Given the Chinchilla Scaling Laws \cite{hoffmann2022training} and the rate at which SoTA LLMs are expanding, \citet{villalobos2022will} estimate that high-quality English training data will be exhausted at some point between 2026 and 2032: language model expansion is outpacing available natural language production. This suggests that\textemdash without models that use significantly less data than current approaches\textemdash LLMs' performance increases will begin to decelerate substantially in the near future.

However, there is a considerable amount of evidence in the literature \citep[e.g.][etc.]{xu-etal-2021-syntax,wu-etal-2021-infusing,prange-etal-2022-linguistic,sachan-etal-2021-syntax,zhou-etal-2020-limit,zhang2020semantics,zhang-etal-2022-extracting} indicating that linguistically-informed LMs\textemdash models whose inputs and/or architectures are augmented by linguistic knowledge\textemdash can improve performance without consuming more text data. 

In this paper, we argue for the use of LMs over logical forms (LFLMs): LMs that take as input semantic representations, rather than text. In particular, we posit the following hypothesis:

\begin{figure}[t]
\centering
\includegraphics[width=65mm, height=36mm]{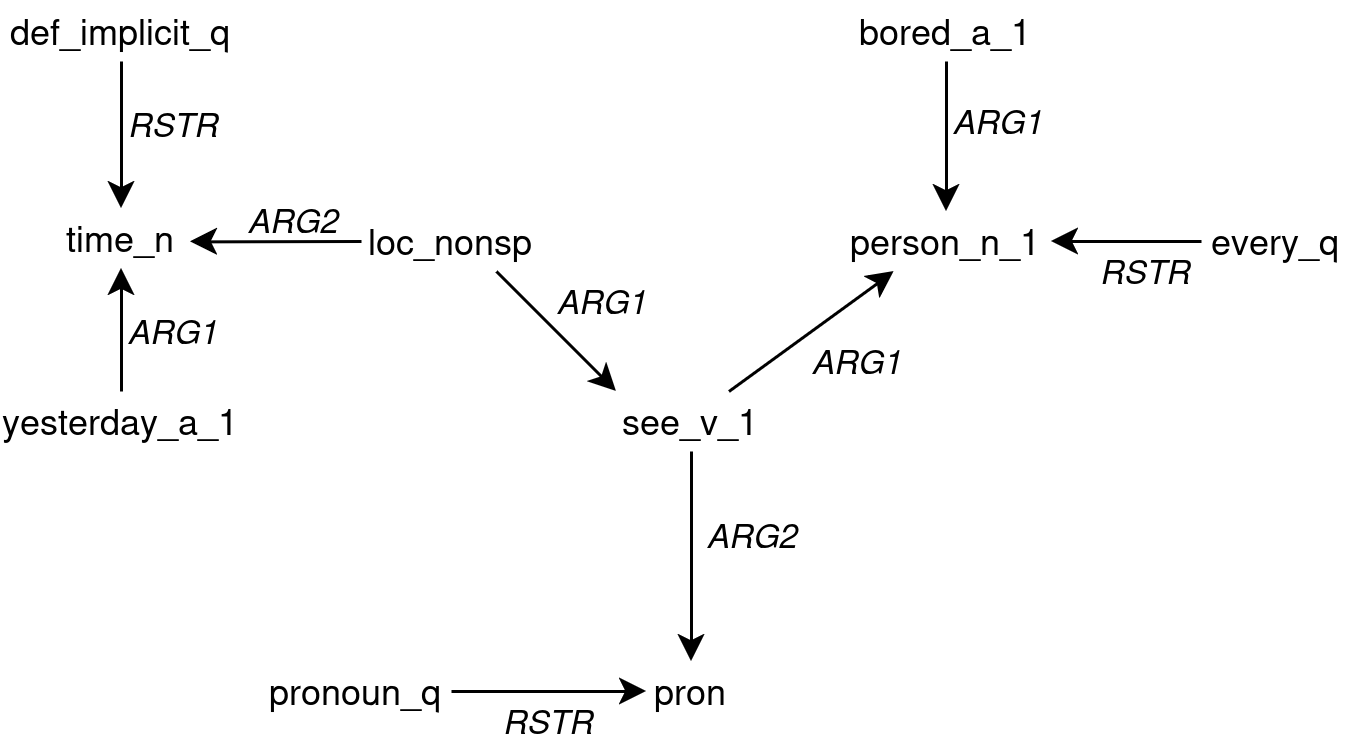}
\caption{DMRS representation of the sentence ``\textit{every bored person saw her yesterday}.''}
\label{fig_dmrs_ex}
\end{figure}

\begin{hypothesis}[The Linguistic-Knowledge Catalysis]
\label{hyp1}
The (aspects of) linguistic knowledge incorporated into LFLMs greatly accelerates their learning of elementary linguistic phenomena, in turn accelerating the learning of more complex patterns.
\end{hypothesis}

One relevant corollary of the \refhyp is that LFLMs can learn with less data: the linguistic knowledge built into LFLMs facilitates more rapid learning of advanced phenomena.

We argue that the primary advantage of logical forms (as opposed to other types of linguistic knowledge) with respect to language modeling is the de-noising effect conferred by the function-argument structure inherent to such representations. Specifically, the translation of surface text to logical form has an equivalence-classing effect, so that all syntactic paraphrases of the same proposition\textemdash for example, an active sentence and its passive counterpart\textemdash are mapped to the same representation. The benefit to an LM of the de-noising effect resulting from this equivalence-classing is clear: the model does not need to learn to equate periphrastic structures, and so can immediately begin learning co-occurrence relations between predicates.

To make the case empirically for LFLMs, we introduce the \textit{\underline{G}raph-based \underline{Fo}rmal-\underline{L}ogical \underline{D}istributional \underline{S}emantics} (GFoLDS; Section \ref{sec_model}) model as a proof-of-concept of such LMs. GFoLDS is a pretrained, encoder graph transformer \cite{wu2021representing} over structures derived from Dependency Minimal Recursion Semantics \citep[DMRS;][]{copestake-2009-invited} representations (see Figure \ref{fig_dmrs_ex}): directed, labeled acyclic graphs in which nodes correspond to predicates and edge labels denote relations (i.e.\hspace{1mm}semantic roles) between them, in a similar manner to Abstract Meaning Representation \citep[AMR;][]{banarescu-etal-2013-abstract} structures. 

Unlike AMR, which abstracts away from morphosyntactic features such as tense and number, DMRS includes these features (see Figure \ref{fig_dmrs_ex_feats} in the Appendix). Beyond yielding a more faithful representation of linguistic meaning, this additionally has the effect of further de-noising the model’s input by offloading the morphological realization of these features to explicitly annotated labels: GFoLDS does not need to learn the surface patterns corresponding to inflection, as this information is instead explicitly provided through the DMRS representation. For example, the model does not need to understand that the suffix \textit{–s} has the same effect on meaning as other, irregular realizations of pluralization (e.g.\hspace{1mm}$\textit{goose}\Rightarrow\textit{geese}$), because plural nouns are directly labeled as such.

% \begin{figure}[ht]
% \centering
% \includegraphics[width=65mm, height=36mm]{figures/dmrs_ex.png}
% \caption{DMRS representation of the sentence ``\textit{every bored person saw her yesterday}.''}
% \label{fig_dmrs_ex}
% \end{figure}

The contributions of this work are threefold: firstly, we provide experimental support towards the validity of the \refhyp, demonstrating that\textemdash from the start of pretraining\textemdash GFoLDS achieves near-peak performance on tasks designed to evaluate its elementary linguistic knowledge, and that this translates to more rapid learning of complex phenomena (Section \ref{sec_hyp1}). 

% The objectives of this work are threefold: firstly, we aim to provide experimental support towards the validity of the \refhyp. This is achieved in Section \ref{sec_hyp1}, where we demonstrate that\textemdash from the start of pretraining\textemdash GFoLDS achieves near-peak performance on tasks designed to evaluate its elementary linguistic knowledge, and that this translates to more rapid learning of complex phenomena. 

Secondly, we demonstrate the viability of pretrained LFLMs in Section \ref{sec_experiments}, by comparing the performance of GFoLDS to that of BERT \citep[trained on $\sim$6.5 times more data than GFoLDS;][]{devlin-etal-2019-bert} on a range of downstream tasks. Although the actual BERT models outperform GFoLDS, our model outperforms\textemdash by a wide margin\textemdash BERT models pretrained on the same data as GFoLDS on all benchmarks, indicating that LFLMs can learn useful representations with much less data than their textual counterparts.

Thirdly, we establish that LFLMs have the potential to compete with textual LLMs at scale: in Section \ref{sec_scalability}, we present evidence indicating that GFoLDS is likely to scale with respect to parameter count and pretraining dataset size.

We make all code for the GFoLDS model and the experiments conducted in this paper available on GitHub\footnote{\href{https://github.com/mjs227/GFoLDS}{https://github.com/mjs227/GFoLDS}}.

\section{Related Work}
\label{sec_background}

As discussed in Section \ref{sec_intro}, there exists a body of research indicating that injecting graph representations of linguistic structures into textual LMs can improve downstream performance. \citet{xu-etal-2021-syntax} fuse dependency parse graphs into pretrainined transformer encoders \citep[e.g. BERT and RoBERTa;][]{liu2019roberta}, and surpass the then-SoTA results on relation classification, entity typing, and question answering tasks, demonstrating the general utility of linguistically-informed LMs. Similarly, \citet{wu-etal-2021-infusing} inject syntactic dependency parse graphs into a pretrained BERT model, yielding then-SoTA results on semantic role labeling and relation extraction tasks. These authors additionally compare the respective impacts of syntactic and semantic representations on performance, and find that semantic representations are more beneficial with respect to downstream performance than syntactic structures.

\citet{prange-etal-2022-linguistic} show that linguistic structures can be incorporated into the input of GPT-2 \citep{gpt2} to improve next word prediction accuracy and entropy. Of particular relevance is their finding that Elementary Dependency Structures \citep[EDS;][]{oepen2006discriminant}\textemdash a semantic framework that is related to DMRS\textemdash yield greater performance improvements than syntactic (or other semantic) representations.

All of the models described thus far in this section are hybrid architectures that merge textual and graph representations. Furthermore, in these approaches, the textual component of the model is initialized from a pretrained LM such as BERT, RoBERTa, or GPT-2. This contrasts with the GFoLDS model, which takes only graph representations as input and is pretrained from scratch. 

To the best of our knowledge, Functional Distributional Semantics at Scale \citep[FDSAS;][]{lo-etal-2023-functional} represents the only\footnote{Aside from its predecessor: Functional Distributional Semantics \cite{emerson2018functional}.} extant LFLM aside from GFoLDS. FDSAS is a variational autoencoder over DMRS graphs that learns probabilistic truth-conditional functions for each predicate (node label). FDSAS is not, however, a transformer and so does not benefit from the flexibility afforded by such architectures: it is unclear how to scale the model to yield a deeper architecture, and it does not generate usable hidden states that can be passed to (for example) a classification head for downstream tasks. Additionally, its rigorous formal-semantic foundations give rise to a degree of inflexibility in this model: for example, \citet{lo-etal-2023-functional} discarded prepositions, quantifiers, and modal verbs from FDSAS' training dataset out of necessity. 

\section{GFoLDS}
\label{sec_model}

In this section, we describe the GFoLDS model architecture (Section \ref{sec_model_sub_arch}), preprocessing steps that we employed on its DMRS graph inputs (Section \ref{sec_model_sub_preproc}), and the model's pretraining procedure (Section \ref{sec_model_sub_pting}). 

\subsection{Architecture}
\label{sec_model_sub_arch}

\begin{figure}[t]
\centering
\includegraphics[width=75mm, height=35mm]{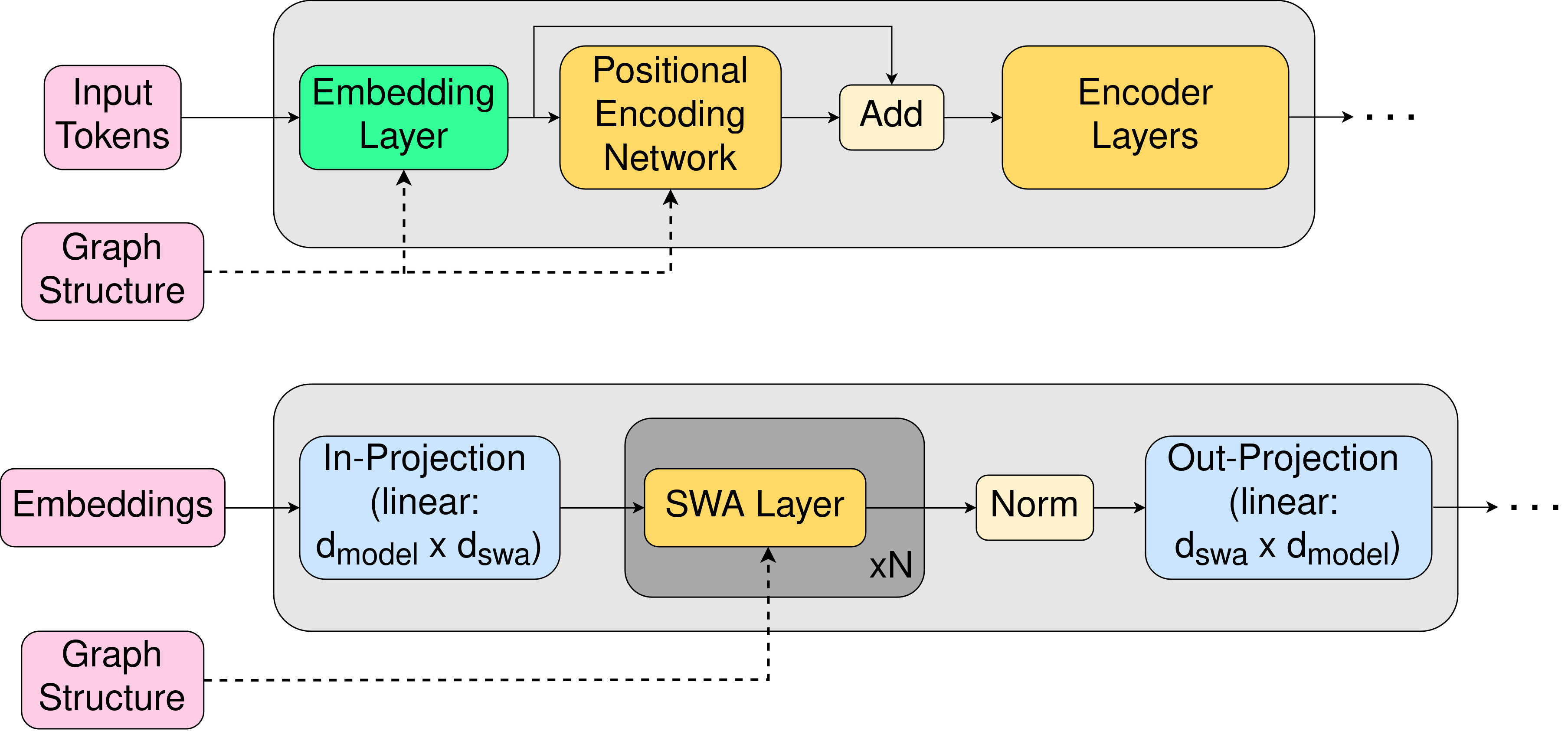}
\caption{Top-level architecture of the GFoLDS model (top) and positional encoding network (bottom).}
\label{fig_gfolds_overall}
\end{figure}

The GFoLDS model is a variant of the graph transformer paradigm \cite{wu2021representing}, which was originally introduced for molecule graph classification. A graph transformer consists of a graph neural network (GNN) that encodes local neighborhood information, whose output is then fed to a permutation-invariant (i.e.\hspace{1mm}without linear positional embeddings) transformer encoder for global message-passing (attention).

Unique to this work is the GNN component of GFoLDS, which consists of an embedding layer and the \textit{positional encoding network} (see Figure \ref{fig_gfolds_overall}). The output of the embedding layer is fed into the positional encoding network, which provides each node with a representation of its local neighborhood in the DMRS graph structure.

% The input to the encoder stack is the sum of the respective outputs of the embedding layer and positional encoding network. This is analogous to (and inspired by) the approach taken by most transformer LLMs \citep[e.g.][etc.]{devlin-etal-2019-bert,brown-etal-2020-gpt3,lewis-etal-2020-bart,raffel2020exploring,dubey2024llama}, in which the input to the encoder stack for a given token $t$ at position $p$ is the sum of the token embedding for $t$ with the positional embedding for $p$.

% The positional encoding network generates a representation of each node that encodes its location relative to that of its neighbors in the graph. As mentioned above, these position-aware node representations are then sent to the transformer encoder: formally, the input to the encoder stack is $E(X,G)+P(E(X,G),G)$, where $E$ denotes the embedding layer, $P$ the positional encoding network, $X$ the input tokens (i.e.\hspace{1mm}node labels), and $G$ the graph structure. The encoder is \textit{not} directly exposed to the graph structure: all nodes are able to attend to any other node(s)\textemdash the degree to which they attend to one another is merely informed by their graph-aware positional encodings.

Let $n_i$ denote the $i^{th}$ node in a graph $G$, and let $F(n_i)$ be its set of DMRS features (person, number, tense, etc.; see Figure \ref{fig_dmrs_ex_feats} in the Appendix). Then the output of the embedding layer $\vec{e_i}=E(X,G)_i$ (where $X$ denotes the node labels of $G$) is the sum of the embedding of the node's label $\mathcal{E}_T(X_i)$ with the normalized sum of the embeddings $\mathcal{E}_F(\phi)$ of each feature $\phi\in F(n_i)$ (Equation \ref{eq_feat}).

\begin{equation}
\label{eq_feat}
   \vec{e_i}=\mathcal{E}_T(X_i)+\text{\textit{Norm}}\left(\sum_{\phi\in F(n_i)}\mathcal{E}_F(\phi)\right)
\end{equation}

\begin{figure}[t]
\centering
\includegraphics[width=65mm, height=21mm]{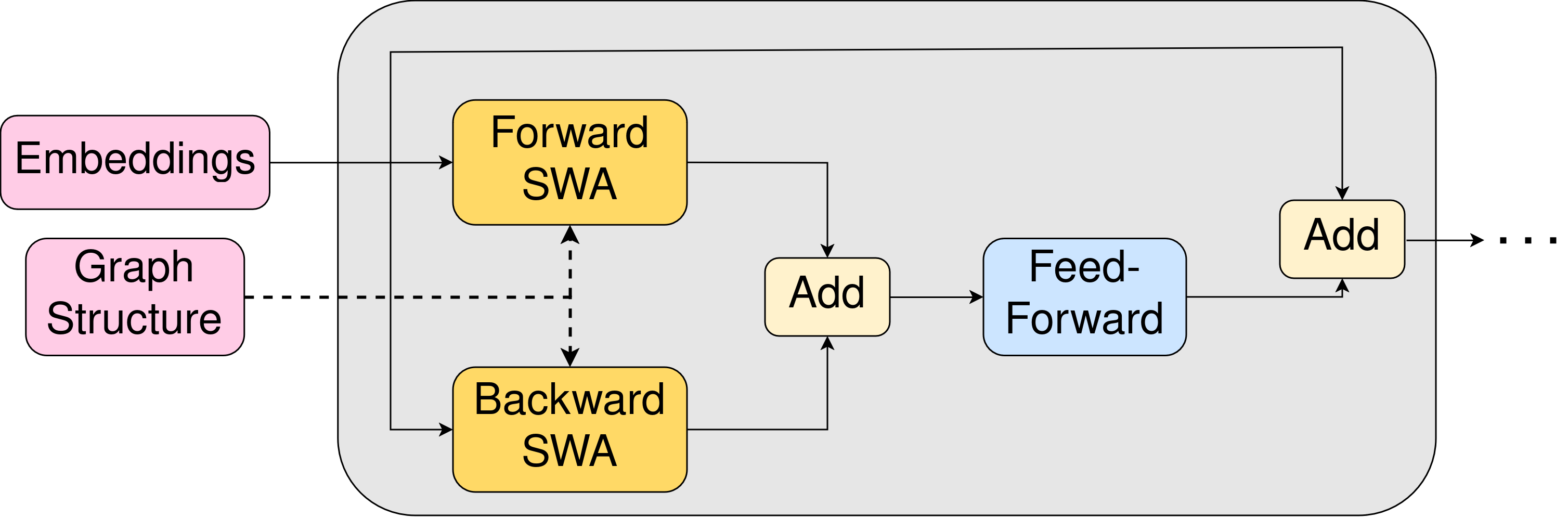}
\caption{Architecture of an SWA layer in the positional encoding network.}
\label{fig_gfolds_swa}
\end{figure}

These summed feature and node/predicate embeddings are then passed to the positional encoding network (see Figure \ref{fig_gfolds_overall}). This module consists of a linear layer (to project the embeddings from $d_{\text{\textit{model}}}$ to $d_{\text{\textit{SWA}}}$) followed by a stack of \textit{step-wise aggregation} (SWA) layers (see Figure \ref{fig_gfolds_swa})\textemdash in which the output of each SWA layer is fed to the subsequent layer\textemdash followed by a second linear projection (to project from $d_{\text{\textit{SWA}}}$ back to $d_{\text{\textit{model}}}$).

In their respective adaptations of the GraphSAGE \citep{hamilton2017inductive} and Graph Convolutional Network \citep[GCN;][]{kipf2017semisupervised} architectures to directed graphs, \citet{xu2018graph2seq} and \citet{tong2020directed} introduce \textit{forward} and \textit{backward} node projection layers, which encode information about incoming and outgoing connections (respectively) for a given node. In a similar fashion, each SWA layer (see Figure \ref{fig_gfolds_swa}) contains a forward (Equation \ref{eq_swa_f}) and a backward SWA block, which encode the nodes\textemdash and the semantic roles thereof\textemdash mapping into and out of a given node.

\begin{equation}
\label{eq_swa_f}
   \vec{f_i}=\text{\textit{Norm}}\left(\sum_{n_k\xrightarrow{\ell}n_i\hspace{0.5mm}\in\hspace{0.5mm}G}W_\ell^{(f)}\vec{h_k}\right)
\end{equation}

For each node $n_i\in G$, its forward representation $\vec{f_i}$ (i.e.\hspace{1mm}the output of the forward SWA block) is the (normalized) sum of $W_\ell^{(f)}\vec{h_k}$ for each node $n_k$ with an edge $n_k\xrightarrow{\ell}n_i$ in the graph structure, where $W_\ell^{(f)}\in\mathbb{R}^{d_{\text{\textit{SWA}}}\times d_{\text{\textit{SWA}}}}$ is the forward \textit{edge projection} linear layer for the edge label $\ell$. This representation of each edge label as a unique projection layer is conceptually similar to the label-specific matrices employed in Relational GCNs \citep{schlichtkrull2018modeling} and \cites{beck-etal-2018-graph} adaptation of the Gated GNN \citep{li2016gated} architecture to labeled graphs.

The backward SWA block is architecturally identical to the forward block, but contains distinct edge projection matrices $W_\ell^{(b)}$ and operates on the transpose $G^T$ of the graph. The respective forward and backward representations of each node $n_i$ are summed together and passed through a two layer, feed-forward block (identical to the feed-forward blocks in the model's encoder layers).

The input to the encoder stack is then the sum of the outputs of the embedding layer and positional encoding network: $E(X,G)+P(E(X,G),G)$. This is analogous to (and inspired by) the approach taken by most transformer LMs \citep[e.g.][etc.]{devlin-etal-2019-bert,brown-etal-2020-gpt3,lewis-etal-2020-bart,raffel2020exploring,dubey2024llama}, where the encoder stack input for token $t$ at position $p$ is the sum of the token embedding for $t$ and the positional embedding for $p$.

The encoder is \textit{not} directly exposed to the graph structure: all nodes are able to attend to any other node(s). The encoder layers in the GFoLDS architecture are similar to those in BERT and \citet{vaswani2017}, with a few key differences with respect to the residual connections and layer normalization. A full description of the GFoLDS architecture is located in Appendix \ref{app_archictecture}.

\subsection{Data Preprocessing}
\label{sec_model_sub_preproc}

GFoLDS' pretraining corpus consisted of $\sim$17.5 million randomly-selected sentences from the November 1, 2023 English Wikipedia dump\footnote{\href{https://huggingface.co/datasets/wikimedia/wikipedia}{https://huggingface.co/datasets/wikimedia/wikipedia}; CC-BY-SA-3.0 license.}, constituting a total of $\sim$508 million words ($\sim$6.5 times smaller than BERT's pretraining corpus). We first used Spacy's SentenceRecognizer\footnote{\href{https://spacy.io/api/sentencerecognizer}{https://spacy.io/api/sentencerecognizer}} pipeline to extract individual sentences from the text. We then used the PyDelphin \citep{Goodman:2019} library with the ACE/ERG \citep{copestake2000} rule-based parser/grammar\footnote{ERG-1214 release: \href{https://github.com/delph-in/erg}{https://github.com/delph-in/erg}} to obtain a DMRS representation of each sentence, before preprocessing the resulting DMRS structures to yield GFoLDS input graphs. The ACE/ERG parser was able to parse $\sim$84\% of the data, for a total of $\sim$14.6 million DMRS-derived graphs.

It was not feasible to tokenize named entities (CARGs) and out-of-vocabulary (OOV) items in the same manner as for the in-vocabulary DMRS predicates\textemdash i.e.\hspace{1mm}by simply assigning an integer to each unique predicate string in the vocabulary. We therefore replaced all CARGs and OOV terms with the [MASK] token. These [MASK] tokens are \textit{not} targets for prediction during the pretraining procedure (as they are OOV, so there is no possible target token): the goal is instead to have the model represent the OOV item with the closest in-vocabulary token, based on the context in which the OOV item appears. The removal of CARGs and OOV items is a stop-gap measure, and remains an open problem and barrier to the performance of the GFoLDS model. We defer the incorporation of CARGs and OOV items into the model's input structures to future work (see Section \ref{sec_limitations_sub_gfolds}).

Further details on the preprocessing procedures that we employed are located in Appendix \ref{app_preproc}.

\subsection{Pretraining}
\label{sec_model_sub_pting}

We pretrained GFoLDS with the \textit{masked-node modeling} (MNM) objective, which is analogous to the MLM objective used to pretrain encoder transformer LMs. The model trained for four epochs, with a total training time of $\sim$102 hours ($\sim$25.5 hours per epoch) on a single NVIDIA A100 GPU. Further details on our pretraining procedure and hyperparameters are located in Appendix \ref{app_pretraining_sub_gfolds}.

We employed a GFoLDS model with two SWA layers and ten encoder layers (eight attention heads each), and set $d_{\text{\textit{SWA}}}=d_{\text{\textit{model}}}=1024$. The MNM prediction head that we used is identical to BERT's MLM prediction head (aside from the difference in vocabulary size). This yields a total of $\sim$174 million parameters: for comparison, $\text{BERT}_\text{base}$ (12 encoder layers, $d_{\text{\textit{model}}}$ = 768) and $\text{BERT}_\text{large}$ (24 encoder layers, $d_{\text{\textit{model}}}$ = 1024) have $\sim$110 million and $\sim$335 million parameters, respectively.

\section{Evaluating the \refhyp}
\label{sec_hyp1}

This section is dedicated to an investigation of the validity of the \refhyp. The hypothesis can be broken down into two distinct claims: (i) that the (aspects of) linguistic knowledge incorporated into LFLMs greatly accelerates their learning of elementary linguistic phenomena; and (ii) that this in turn accelerates the learning of more complex patterns.

We therefore divided this experiment into two parts, which respectively probe the model's knowledge of elementary and complex linguistic phenomena. Due to the claim made in the \refhyp that linguistically-informed LMs' learning of complex patterns is \textit{accelerated}, we evaluate the model at regular intervals throughout pretraining, in order to measure the rate at which it is learning. 

\subsection{Comparison Models}
\label{sec_hyp1_sub_comp_models}

As a baseline, we pretrained $\text{BERT}_{\text{base}}$ and $\text{BERT}_{\text{large}}$ (uncased) models from scratch on the same dataset as GFoLDS (the surface sentences), for the same number of epochs (four). 

Recall that the ACE/ERG parser was only able to parse $\sim$84\% of the sentences in GFoLDS' pretraining dataset (see Section \ref{sec_model_sub_preproc}). If GFoLDS were able to match the performance of BERT with $\sim$84\% of the pretraining data (for example), then\textemdash given the ACE/ERG parser's $\sim$84\% successful parse rate\textemdash this model would provide no practical benefit over its textual counterparts. We therefore chose to pretrain the BERT comparison models (BERT-C) on GFoLDS' entire pretraining dataset, rather than the parsable subset: the BERT-C models were pretrained with $\sim$1.19 times more data than GFoLDS.

We evaluated a variety of different pretraining hyperparameter configurations for BERT-C in order to yield the most rigorous comparison: a full description of these hyperparameter configurations is located in Appendix \ref{app_pretraining_sub_bert}.

\subsection{Experimental Setup}
\label{sec_hyp1_sub_setup}

This experiment consisted of two tasks that were designed to evaluate knowledge of elementary linguistic phenomena, and one to evaluate knowledge of more complex patterns: we evaluated GFoLDS and the BERT comparison models on the three tasks at twenty evenly-spaced intervals per pretraining epoch, for a total of eighty points of comparison.

Note that it is difficult\textemdash if not impossible\textemdash to define ``elementary'' and ``complex'' linguistic phenomena in absolute terms. In this paper, we consider these terms in a relative sense: the elementary tasks (Section \ref{sec_hyp1_sub_setup_sub_elem}) are undoubtedly \textit{more elementary} than the complex task (Section \ref{sec_hyp1_sub_setup_sub_relpron}), in that they do not require as much (if any) world knowledge, and deal entirely with awareness of basic linguistic categories. Conversely, this entails that our complex task is complex \textit{relative to} our elementary tasks. 

Assuming that the \refhyp holds, we should expect to see GFoLDS outperform the BERT-C models on the complex task throughout the pretraining process, as\textemdash according to the hypothesis\textemdash GFoLDS is able to learn complex patterns faster than textual LMs. 

On the elementary tasks, we again expect GFoLDS to outperform BERT-C, but also that GFoLDS' performance will improve substantially faster than it does on the complex tasks: the \refhyp predicts that an LFLM's accelerated learning of elementary phenomena catalyzes its learning of complex patterns, so its learning of the former should therefore accelerate at a faster rate than that of the latter.

% On the elementary tasks, we expect GFoLDS to outperform BERT-C, but also that GFoLDS's performance will \textit{not} improve significantly across pretraining, while BERT-C's performance steadily increases: the \refhyp asserts that an LFLM is already equipped with elementary linguistic knowledge, so its performance at the first checkpoint should be near (or above) that of a textual model such as BERT-C at the \textit{last} checkpoint. 

\subsubsection{Complex Task}
\label{sec_hyp1_sub_setup_sub_relpron}
The complex task in this experiment is the RELPRON \cite{rimell-etal-2016-relpron} dataset. This dataset consists of \textit{terms} (nouns), each paired with a hypernym and up to ten \textit{properties}: relative clauses that restrict that hypernym (see Table \ref{table_relpron_ex} in the Appendix). The development set consists of 65 terms and 518 properties ($\sim$8 properties per term on average), and the test set contains 73 terms and 569 properties ($\sim$7.8 per term). The task is to retrieve the properties that apply to each term, without including those that do not: the evaluation metric is Mean Average Precision (MAP) score. 

To evaluate the models, we constructed templates out of each (term, hypernym, property) triple: for example, the triple (\textit{telescope}, \textit{device}, \textit{astronomers use}) yields the template ``\textit{a device that astronomers use is a \underline{telescope}}''. We then replaced the target term with the [MASK] token (e.g.\hspace{1mm}``\textit{a device that astronomers use is a} [MASK]''): the probability assigned to a given term under the masked distribution is taken as proportional to the probability that the property applies to that term.

% To evaluate the models on RELPRON, we constructed templates out of each (term, hypernym, property) triple: for example, the triple (\textit{telescope}, \textit{device}, \textit{astronomers use}) yields the template ``\textit{a device that astronomers use is a \underline{telescope}}''. We then replaced the target term with the [MASK] token: the probability assigned to a given term under the masked distribution is taken as proportional to the probability that the property applies to that term.

\begin{figure*}[t]
\centering
\includegraphics[width=145mm, height=38mm]{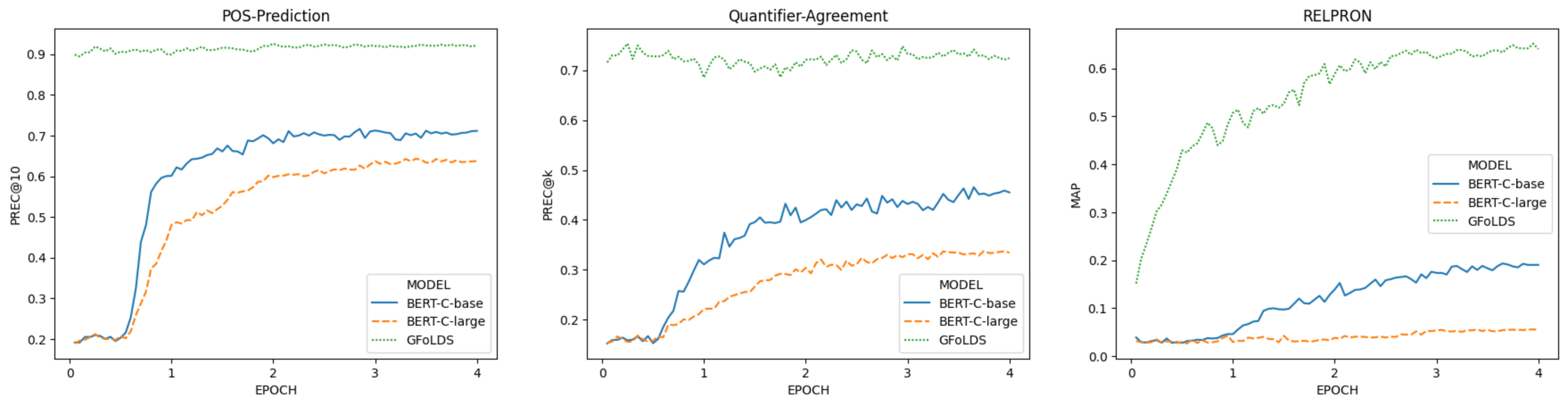}
\caption{Precision scores on the two elementary tasks (left and center) and MAP scores on the RELPRON test set (right) across the 80 evenly-spaced training snapshots for GFoLDS (green) and the $\text{BERT}_\text{base}$/$\text{BERT}_\text{large}$ comparison models (blue/orange, respectively).}
\label{fig_analysis_res}
\end{figure*}

As discussed in Section \ref{sec_model_sub_preproc}, CARGs and OOV items are masked in the input graphs: given that each template only contains four content words (the hypernym, verb, relative clause subject/object, and the target term), the GFoLDS model is effectively blind to (at least) one third of the context in templates that contain OOV items or CARGs. We therefore  discarded all examples containing CARG-bearing predicates or OOV items for evaluation. This resulting subset of the test split (the ``RELPRON-No-UNK/NE'' column of Table \ref{table_downstream_res}) contains 63 terms and 421 properties, for a total of $\sim$6.68 properties per term on average.

Due to the small size of the dataset (and the lack of a training split), the frozen, pretrained models were used to obtain token probabilities for property ranking on this task.

\subsubsection{Elementary Tasks}
\label{sec_hyp1_sub_setup_sub_elem}

\paragraph{POS-Prediction.} The first elementary task evaluates the LMs' ability to model the distribution of parts-of-speech: this is a commonly-employed probing task used to assess LMs' elementary linguistic competence \citep{waldis2024holmes}. We evaluated the models on 200 sentences (not drawn from their pretraining data) from English Wikipedia. For each sentence, we masked a single word and recorded its POS: we then extracted the models' probability distribution over the masked word of each sentence, and recorded their precision at ten with respect to the masked word's part-of-speech. For example, if the masked word was a noun, and nine of the model's top ten most-likely predictions were nouns, then the model received a score of 0.9 for that instance.

\paragraph{Quantifier-Agreement.} The second elementary task evaluates the models' knowledge of quantifier number agreement\textemdash whether a quantifier restricts a singular or plural noun (or both)\textemdash across 179 sentences drawn from English Wikipedia (not used to pretrain the models). Although less common than POS prediction, quantifier agreement tasks have also been employed as a metric of language models' elementary lingustic competence \citep[see e.g.][]{huebner-etal-2021-babyberta,waldis2024holmes}.

As in the POS-prediction task above, we masked a single quantifier per sentence, then extracted the probability distribution over the masked quantifier of each sentence to compute precision at $k$. The value of $k$ varied as a function of the number of the noun in the masked quantifier's restriction, as there were more quantifiers that restrict singular nouns ($k=14$) than plural nouns ($k=13$). 

Additional details on the setup and evaluation of the elementary tasks are given in Appendix \ref{app_elementary}.

\subsection{Results}
\label{sec_hyp1_sub_results}

The results of this experiment (Figure \ref{fig_analysis_res}) conform almost exactly to the behavior predicted by the \refhyp: the performance of GFoLDS on the elementary tasks remains relatively constant throughout pretraining, because the model starts near peak performance from the first checkpoint\textemdash this indicates that its learning of elementary patterns was complete within 5\% of the first epoch. While the performance of the BERT models steadily increases on these tasks, it does so at a much lower rate. 

On the RELPRON test set, GFoLDS begins improving immediately, $\text{BERT-C}_\text{large}$ does not improve substantially, and the performance of $\text{BERT-C}_\text{base}$ doesn't begin to meaningfully increase until the latter half of the first epoch (and at a lower rate than that of GFoLDS): the point at which it began to improve on the elementary tasks.

These results constitute strong evidence towards the \refhyp. It is clear from this experiment that GFoLDS can model the elementary phenomena almost from the onset (and retains this ability throughout pretraining) and GFoLDS' performance on the RELPRON test set suggests that this accelerated learning of elementary phenomena translates to more rapid learning of more complex patterns. 

\section{Downstream Tasks}
\label{sec_experiments}

In order to demonstrate the viability of LFLMs, we evaluated GFoLDS on a series of downstream tasks: our primary objective was to demonstrate that language models over logical forms are able to learn from less data than their textual counterparts. 

To that end, we compared GFoLDS to the BERT comparison (BERT-C) models pretrained on the same data (see Section \ref{sec_hyp1_sub_comp_models}), in order to show that GFoLDS is able to outperform a textual LM when pretrained on the \textit{same} amount of data: it then follows that GFoLDS would be able to perform as well as a textual LM with \textit{less} pretraining data.

We additionally compared GFoLDS to the original $\text{BERT}_{\text{base}}$ and $\text{BERT}_{\text{large}}$ (uncased) models, both of which are pretrained on roughly 6.5 times more data than GFoLDS.

\subsection{Datasets}
\label{sec_experiments_sub_datasets}
We evaluated the models on four downstream tasks: the RELPRON task described in Section \ref{sec_hyp1_sub_setup_sub_relpron}, SNLI \cite{bowman2015large}, the MegaVeridicality V2.1 factuality dataset \cite{white-etal-2018-lexicosyntactic}, and the \citet{mcrae-etal-2005-semantic} property inference dataset.

\subsubsection{SNLI}
\label{sec_experiments_sub_datasets_sub_snli}

% The SNLI \cite{bowman2015large} dataset consists of 550,152 training examples and 10,000 development/test examples (each). The ACE/ERG parser was able to parse $\sim$88\% of these premise/hypothesis pairs, for a total of 8,813 DMRS graph pairs in the test split (3,053 entailment; 2,887 contradiction; 2,873 neutral) and 487,590 in the training split (163,670 entailment; 162,476 contradiction; 161,444 neutral).

All five models were fine-tuned end-to-end on SNLI with a two-layer, feed-forward MLP classifier over their mean-pooled token/node embeddings. The details of our SNLI fine-tuning hyperparameters are located in Appendix \ref{app_finetuning}.

\paragraph{Graph Representations.} The GFoLDS model described in Section \ref{sec_model} cannot process multiple sentences. While this obviously represents a serious general limitation of the model (discussed further in Section \ref{sec_limitations}), it also presents a more immediate complication in terms of the SNLI dataset, whose examples are given as $(\text{premise},\text{hypothesis})$ pairs. 

\begin{figure}[t]
\centering
\includegraphics[width=65mm, height=58mm]{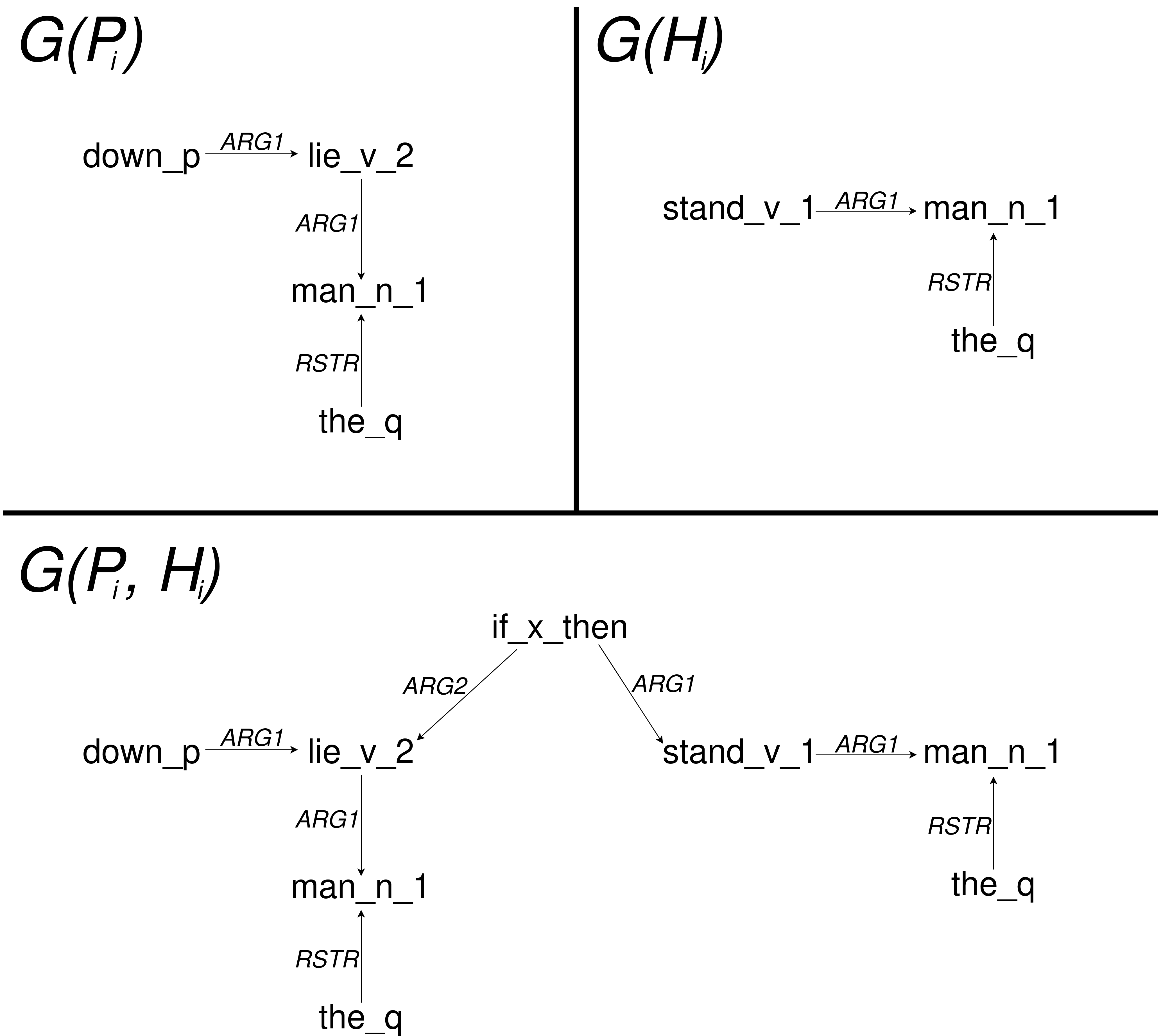}
\caption{Illustration of the derivation of $G(P_i,H_i)$ (bottom) from $G(P_i)$ (top left) and $G(H_i)$ (top right) for the contradiction example (\textit{the man is lying down, the man is standing}).}
\label{fig_if_x_then_ex}
\end{figure}

\begin{table*}[t]
\centering
\scalebox{0.8}{\begin{tabular}{l|lllll}
& \textbf{RELPRON-All} & \textbf{RELPRON-No-UNK/NE} & \textbf{SNLI} & \textbf{MegaVeridicality V2.1} & \textbf{\citet{mcrae-etal-2005-semantic}} \\
& \textbf{(MAP)} & \textbf{(MAP)} & \textbf{(Accuracy)} & \textbf{(Accuracy)} & \textbf{(Spearman $\rho$)} \\
\hline
GFoLDS & \textemdash & 0.651 & 81.0\% & 81.3\% & 0.205 \\
$\text{BERT-C}_{\text{base}}$ & 0.147 & 0.193 & 79.9\% & 78.1\% & 0.167 \\
$\text{BERT-C}_{\text{large}}$ & 0.047 & 0.056 & 62.0\% & 76.2\% & 0.134 \\
$\text{BERT}_{\text{base}}$ & 0.667 & 0.690 & 90.7\% & 84.2\% & 0.247 \\
$\text{BERT}_{\text{large}}$ & 0.768 & 0.769 & 91.1\% & 85.6\% & 0.241 \\
FDSAS & 0.580 & \textemdash & \textemdash & \textemdash & \textemdash
\end{tabular}}
\caption{Results for GFoLDS and the four BERT models on downstream tasks. $\text{BERT-C}_\text{base/large}$ indicates the BERT models pretrained on the same data as GFoLDS (see Section \ref{sec_hyp1_sub_comp_models}).}
\label{table_downstream_res}
\end{table*}

% \begin{table*}[t]
% \centering
% \scalebox{0.825}{\begin{tabular}{l|llllll}
% & GFoLDS & $\text{BERT}_{\text{large}}$ & $\text{BERT}_{\text{base}}$ & $\text{BERT-C}_{\text{large}}$ & $\text{BERT-C}_{\text{base}}$ & FDSAS \\
% \hline
% \textbf{RELPRON-All (MAP)} & \textemdash & 0.768 & 0.667 & 0.047 & 0.174 & 0.580 \\
% \textbf{RELPRON-No-UNK/NE (MAP)} & 0.651 & 0.769 & 0.690 & 0.056 & 0.193 & \textemdash \\
% \textbf{SNLI (Accuracy)} & 81.0\% & 91.1\% & 90.7\% & 62.0\% & 79.9\% & \textemdash \\
% \textbf{MegaVeridicality V2.1 (Accuracy)} & 81.3\% & 85.6\% & 84.2\% & 76.2\% & 78.1\% & \textemdash \\
% \textbf{\citet{mcrae-etal-2005-semantic} (Spearman $\rho$)} & 0.205 & 0.241 & 0.247 & 0.134 & 0.167 & \textemdash \\
% \end{tabular}}
% \caption{Results for GFoLDS and the four BERT models on downstream tasks.}
% \label{table_downstream_res}
% \end{table*}

To overcome this obstacle, we construct a single graph $G(P_i,H_i)$ from each premise ($G(P_i)$) and hypothesis ($G(H_i)$) pair. $G(P_i,H_i)$ is derived from the disjoint union $G(P_i)\oplus G(H_i)$ by adding the node \textit{if\_x\_then}, and then inserting edges\footnote{The DMRS \textit{htop} node is typically the (node corresponding to the) main verb of the sentence from which $G$ is derived: see \citet{copestake2005minimal}.} $\text{\textit{if\_x\_then}}\xrightarrow{\text{\textit{ARG1}}}\text{\textit{htop}}(G(H_i))$ and $\text{\textit{if\_x\_then}}\xrightarrow{\text{\textit{ARG2}}}\text{\textit{htop}}(G(P_i))$, as in Figure \ref{fig_if_x_then_ex}. 

% We defer readers to \citet{copestake2005minimal} for a discussion of DMRS \textit{htop} nodes: here, it suffices to state that $\text{\textit{htop}}(G)$ is typically the (node corresponding to the) main verb of the sentence from which $G$ is derived.

\subsubsection{MegaVeridicality V2.1}
\label{sec_experiments_sub_datasets_sub_factuality}

The objective in this task is to determine whether the event denoted by a subordinate clause is true, given the context of the matrix clause (see Table \ref{table_factuality_ex} in the Appendix). The MegaVeridicality V2.1 dataset consists of 5,026 examples, each with ten annotations of the factuality of the event: for this experiment, we converted the data into a binary classification task via majority voting of the annotated labels (see Appendix \ref{app_factuality}).

As with RELPRON (see Section \ref{sec_hyp1_sub_setup_sub_relpron}), we removed all examples containing OOV items and/or CARG-bearing predicates for this task, due to the lower amount of content words per example in MegaVeridicality V2.1 in comparison to SNLI. This left 3,126 remaining examples ($\sim$62\%), of which we withheld 20\% as a test set.

% As with SNLI (Section \ref{sec_experiments_sub_datasets_sub_snli}), we fine-tuned all five models end-to-end with a two-layer (binary) classifier over their mean-pooled embeddings. The details of our fine-tuning hyperparameter configurations for this task are located in Appendix \ref{app_factuality_sub_ft}.

We fine-tuned all five models end-to-end for eight epochs with a two-layer (binary) classifier head over their mean-pooled embeddings, using a learn rate of $10^{-6}$, a weight decay value of $10^{-5}$, and a batch size of 16.

\subsubsection{The \citet{mcrae-etal-2005-semantic} Dataset}
\label{sec_experiments_sub_datasets_sub_propinf}

To evaluate the quality of GFoLDS' embeddings in comparison to BERT, we evaluated the models on a property inference task using the \citet{mcrae-etal-2005-semantic} feature norm database. This database consists of a set of 541 \textit{concepts} (words) $W$ and 2,526 \textit{features} $F$, where each concept $w\in W$ is assigned a feature vector $f(w)\in\mathbb{R}^{|F|}$: the value of $f(w)_Q$ is the value of the feature $Q$ for the word $w$ (see Table \ref{table_mcrae_ex} in the Appendix). 

Following \citet{rosenfeld-2022-analysis}, we created ten random folds consisting of 50 concepts each from the dataset. The concepts within each fold represent the set $U$ of \textit{unknown} words, and those outside of the fold represent the set $K=W-U$ of \textit{known} words. For each $u\in U$, its feature vector $f(u)$ is withheld: the task is to reconstruct $f(u)$.

As the goal of this experiment was to evaluate the quality of their embeddings, the models were not fine-tuned for this task: again following \citet{rosenfeld-2022-analysis}, we used the Modified Adsorption \cite{talukdar-2009-regularized} graph label-propagation algorithm to estimate properties, using the cosine distance between two words' embeddings to weight the graph edge between them. 

The evaluation metric was Spearman's $\rho$: we averaged over all scores for each unknown word in each fold to yield each model's final score.

\subsection{Results}
\label{sec_experiments_sub_results}

The results of these experiments are given in Table \ref{table_downstream_res}. Although the original BERT models outperform GFoLDS, the BERT comparison models trained on the same data as our model (BERT-C; see Section \ref{sec_hyp1_sub_comp_models}) both lag far behind GFoLDS on all four benchmarks, demonstrating across a wide range of downstream tasks that our model is able to learn useful representations with less data than its textual counterparts.

The pretrained FDSAS model is not publicly available, so we were unable to evaluate it on the RELPRON No-UNK/NE subset to obtain a direct comparison to GFoLDS. However, based on the differences in MAP scores from the full dataset to the No-UNK/NE subset for $\text{BERT}_{\text{base}}$/$\text{BERT-C}_{\text{base}}$ and $\text{BERT}_{\text{large}}$/$\text{BERT-C}_{\text{large}}$ (+0.023/+0.019 and +0.001/+0.009, respectively), it is likely that GFoLDS would outperform FDSAS (0.580) on that data.

\section{Scalability}
\label{sec_scalability}

While the results of Section \ref{sec_experiments} show that GFoLDS outperforms textual models trained on similar amounts of data, this model is still outperformed by the original BERT models. It is therefore crucial to establish the scalability of GFoLDS: the degree to which we would expect its downstream performance to scale if it were larger and/or pretrained on more data. To that end, we applied the techniques of \citet{muennighoff2024scaling} to GFoLDS, to determine the degree to which our model is under- or over-parameterized\textemdash and therefore, by the scaling laws established in \citet{muennighoff2024scaling}, over- or under-trained.

As loss is not an exact predictor of downstream performance \citep{shin-etal-2022-effect,tay2022scale,xia-etal-2023-training}, we follow \citet{hoffmann2022training} and evaluate the impact of pretraining token count on a validation task as well. We again used the RELPRON dataset for this purpose, as SNLI introduces potential confounding factors in the form of the fine-tuning procedure.

\subsection{Experimental Setup}
\label{sec_scalability_sub_setup}

We pretrained five GFoLDS models on 50\%, 25\%, 12.5\%, 6.25\%, and 3.125\% of the pretraining data used in Section \ref{sec_model_sub_pting}. Following \citet{muennighoff2024scaling}, we ensured that the iterations pretrained on less data always use a randomly-selected subset of the dataset used in those with more data. 

Aside from the differing number of pretraining tokens, all models were pretrained using the same procedure and hyperparameters as described in Section \ref{sec_model_sub_pting}, as the focus of this experiment was the effect of pretraining tokens with a fixed parameter count.

\subsection{Results}
\label{sec_scalability_sub_results}

\begin{figure}[t]
\centering
\includegraphics[width=75mm, height=29mm]{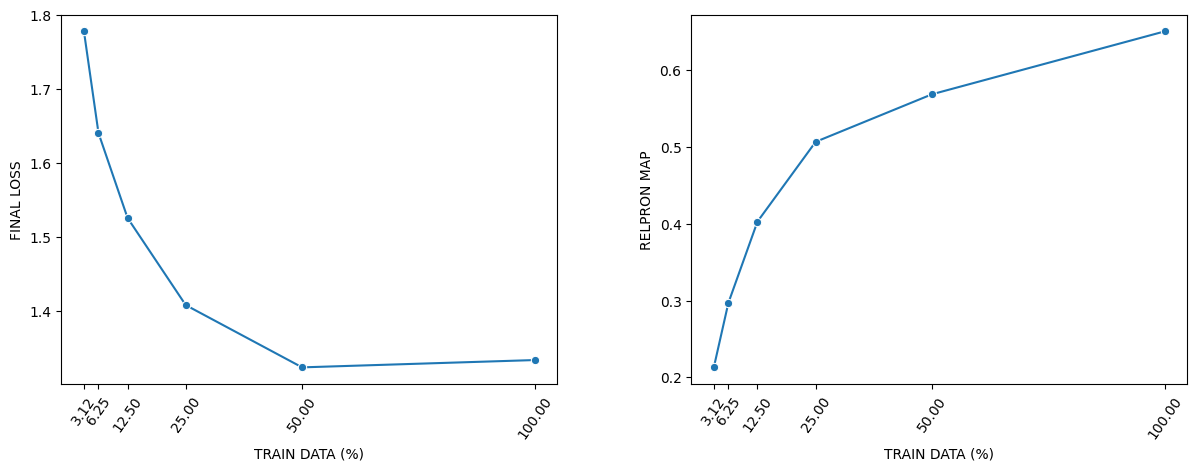}
\caption{Final pretraining loss (left) and RELPRON MAP (right; No-UNK/NE test split). The 100\% run denotes GFoLDS pretrained on the entire dataset.}
\label{fig_scalability_res}
\end{figure}

Figure \ref{fig_scalability_res} shows that final pretraining loss consistently decreases as the number of pretraining tokens increases from 3.125\% to 50\% of the data. After this point, the final loss value plateaus: the 50\% run (1.3232) finishes with a slightly \textit{lower} cross-entropy loss than the actual (100\%) run (1.3331), although this minor difference is likely explained by the noise introduced by random initialization of the models' parameters. 

In Appendix \ref{app_scalability}, we prove that\textemdash assuming that an analogue of the \citet{muennighoff2024scaling} scaling laws holds for the GFoLDS architecture\textemdash it can only be the case that the final loss for the 100\% run is (roughly) equal to that for the 50\% run if GFoLDS is underparameterized for both the 100\% run \textit{and} the 50\% run. This is to say that GFoLDS requires much less pretraining data per parameter than textual models: the Chinchilla Scaling Laws \cite{hoffmann2022training} predict that a textual LLM with the same parameter count ($\sim$174 million) as GFoLDS necessitates $\sim$5 billion pretraining tokens\textemdash roughly twenty times more than the $\sim$254 million tokens for which GFoLDS is overparameterized.

We also observe improvement of +0.082 in RELPRON MAP score from the 50\% (0.569) to the 100\% (0.651) run\textemdash higher than the +0.062 increase from the 25\% (0.507) to the 50\% run.

The results of this experiment lead to the conclusion that GFoLDS is likely scalable in terms of model size and pretraining data. The application of the \citet{muennighoff2024scaling} scaling laws to GFoLDS indicates that increasing the parameter count will decrease final pretraining loss, while its run-over-run performance on the RELPRON test set suggests that larger pretraining datasets will lead to corresponding increases in the model's downstream performance.

\section{Conclusion}
\label{sec_conclusion}

In Section \ref{sec_hyp1}, we provided direct evidence in support of the \refhyp: critically, this finding indicates that \textit{LFLMs can learn with less data than textual models}. This is supported by our findings in Section \ref{sec_scalability}, which indicate GFoLDS is \textit{under}parameterized with half of its pretraining data.

To the best of our knowledge, the experiments in Section \ref{sec_experiments} represent the first time that a language model pretrained solely over logical forms has been evaluated on a wide range of downstream tasks. GFoLDS' capacity to be applied to tasks ranging from RELPRON to SNLI\textemdash and consistently outperform BERT models pretrained on the same amount of data\textemdash demonstrates this model's versatility. These results therefore represent a significant step towards demonstrating the practical viability of LFLMs.

Although much work remains to be done before LFLMs can reach the same level of performance and utility as their textual counterparts (see Section \ref{sec_limitations}), the results of Sections \ref{sec_hyp1}-\ref{sec_scalability} suggest that such models present an exciting avenue for continuing the improvement of language models at a more sustainable rate of data consumption than LLMs over text.

\section{Limitations}
\label{sec_limitations}

This section is divided into two parts: Section \ref{sec_limitations_sub_gfolds} overviews the limitations of the current GFoLDS pipeline laid out in Section \ref{sec_model}, while Section \ref{sec_limitations_sub_experiments} discusses the limitations of the experiments conducted in Sections \ref{sec_hyp1}-\ref{sec_scalability}.

\subsection{Limitations of the Current Approach}
\label{sec_limitations_sub_gfolds}

\paragraph{Tokenization.}
As discussed in Section \ref{sec_model_sub_preproc}, we replaced all OOV node labels with the [MASK] token. Although this preprocessing step was necessary due to the tokenization scheme that we employed, it prevents the model from seeing every word in every input sentence. For the same reason, we removed CARGs (named entities), keeping only the CARG-bearing predicates (nodes): this amounts to, for example, replacing the sentence ``\textit{John went to the park in the spring of 2017}'' with ``[NAMED]\textit{ went to the park in the }[SEASON]\textit{ of }[YEAR].''

The removal of these items from its DMRS-derived input graphs almost certainly negatively impacts GFoLDS' performance on downstream tasks, and entirely precluded a direct comparison with FDSAS on the RELPRON benchmark (see Section \ref{sec_experiments_sub_results}).  However, this limitation lies purely with the current GFoLDS architecture\textemdash rather than the proposed approach in and of itself\textemdash as we simply tokenized the nodes by assigning a unique integer to each in-vocabulary term (i.e.\hspace{1mm}node label). The ACE/ERG parser can still parse OOV items: the node label of an OOV term is constructed by appending the suffix ``\textit{\_unknown}'' (along with a part-of-speech tag) to the corresponding string in the surface text. 

Future work in this area must involve redesigning the model's approach to node embeddings in order to overcome this limitation, for example by replacing GFoLDS' embedding layer with a small, character-level encoder transformer that is applied to each individual node. As we intended this work to primarily be a proof-of-concept of LFLMs, we left this issue to future research in order to retain focus on that goal. 

\paragraph{Multiple-Sentence Inputs.}
For the SNLI dataset, we were able to overcome GFoLDS' inability to process sequences of multiple sentences by converting each premise/hypothesis pair into a single, connected graph. However, this task-specific remedy cannot be readily generalized to other multiple-sentence NLP benchmarks. In future work, we intend to address this deficiency through sentence-level positional encodings \citep[e.g. rotary embeddings;][]{su2024roformer}: a sequence of sentences represented as the disjoint union of their DMRS graphs, where the positional encoding $\vec{p_i}$ is added (or concatenated) to the embedding of each node in the $i^{th}$ graph in the sequence.

\paragraph{DMRS Parsing.}
Although the ACE/ERG parser that we employed to derive DMRS graphs from natural language enjoys high precision \citep[93.77\%;][]{zamaraeva-gomez-rodriguez-2024-revisiting}, it suffers from relatively low recall: only $\sim$84\% of our pretraining dataset was able to be parsed. This can (and should) be addressed in future work, perhaps through the use of a neural DMRS parser \citep[e.g.][]{buys-blunsom-2017-robust}, which trades lower precision for higher recall than a rule-based parser such as ACE/ERG.

Additionally, the grammar of the ACE/ERG parser (ERG: English Resource Grammar) is solely an \textit{English} grammar, rendering our approach inapplicable to other languages\footnote{Although broad-coverage grammars do exist for some other higher-resource languages \citep[e.g. the Spanish Resource Grammar;][]{zamaraeva-etal-2024-spanish}}.

\paragraph{Model Type.}
The GFoLDS model described in this work is an \textit{encoder} model: in order to be able to perform the same range of tasks as current SoTA LLMs, LFLMs must have generative capabilities. In future research, we intend to adapt GFoLDS to construct a sentence-level, graph-to-graph generative LFLM, by adapting existing work in the domain of molecule graphs on autoregressive graph generation \citep[e.g.][]{bacciu2020edge,goyal2020graphgen,bacciu2021graphgen} that we believe can be readily extended to DMRS, using (for example) a breadth-first search to impose a canonical generation order on the graph nodes during training.

\subsection{Experimental Limitations}
\label{sec_limitations_sub_experiments}

\paragraph{Logical Representations.}
While the GFoLDS model demonstrates that language models over DMRS representations are able to learn with less data than textual models, we did not evaluate the \refhyp using other logical-form representations such as AMR. AMR in particular is a problematic representational format for the GFoLDS model, as this framework utilizes $\sim$100 edge labels \cite{banarescu-etal-2013-abstract} (our DMRS-derived graphs have nine; see Table \ref{table_dmrs_edge_lbl} in the Appendix). Recall from the discussion in Section \ref{sec_model_sub_arch} that each unique edge label corresponds to unique forward/backward edge projection layers within each SWA layer (see also Appendix \ref{app_preproc}): with the architectural configuration of the GFoLDS model used in this work (i.e.\hspace{1mm}two SWA layers with $d_{\text{\textit{SWA}}}=1024$), the additional AMR edge labels would add $\sim$380 million parameters to the positional encoding module (the model in this work has only 174 million \textit{total} parameters).

Although they may require significant architectural modifications, future work should investigate the use of AMR, EDS, and other graph-based semantic representations for language modeling, in order to determine the most effective framework for this application.

\paragraph{Dataset Size.}
Due to computational limitations, we did not investigate the GFoLDS model at scale (i.e.\hspace{1mm}with the same amount of training data as BERT): it required roughly one month to parse the training dataset used in this work into DMRS\footnote{We do not view this as a severe limitation of the proposed approach: DMRS parsing can be achieved in parallel (each parse is independent of the others), and so the speed of parsing scales linearly with the number of CPU cores. With sixteen times as many cores as used in this work, for example, a dataset the size of BERT's pretraining data could be parsed in roughly two weeks.}. Although this does introduce the risk that our proposed method may only show superiority on a small-scale dataset, the results of Section \ref{sec_scalability} indicate that this risk is relatively low: extrapolating from the pattern shown in Figure \ref{fig_scalability_res} (see the discussion in Section \ref{sec_scalability_sub_results}), we would expect GFoLDS to surpass $\text{BERT}_\text{large}$ (the original model) on the RELPRON test set with four times the current training data (still less than that of $\text{BERT}_\text{large}$, which additionally has twice the parameter count).

\paragraph{Model Size.}
We did not investigate the effect of parameter size on the GFoLDS model: these experiments were only performed with a 174 million parameter LFLM.

\section*{Acknowledgments}

Computational resources for these experiments were provided by the Center for Computational Research at the \citet{ccr}.

\bibliography{anthology,custom}

\appendix

\begin{table*}[h]
\centering
\begin{tabular}{|ll|l|l|}
\hline
\textbf{Term} & \textbf{Hypernym} & \textbf{Properties} & \textbf{Corresponding Templates} \\
\hline
\textit{telescope} & \textit{device} & \textit{astronomers use} & ``A device that astronomers use is a telscope'' \\
& & \textit{observatory has} & ``A device that an observatory has is a \\
& & & telescope'' \\
& & \textit{dome houses} & ``A device that a dome houses is a telescope'' \\
& & \textit{observer points} & ``A device that an observer points is a \\
& & & telescope'' \\
& & \textit{has a mirror} & ``A device that has a mirror is a telescope'' \\
& & \textit{uses a lens} & ``A device that uses a lens is a telescope'' \\
& & \textit{detects planets} & ``A device that detects planets is a telescope'' \\
& & \textit{views stars} & ``A device that views stars is a telescope'' \\
& & \textit{tracks the sky} & ``A device that tracks the sky is a telescope'' \\
& & \textit{collects light} & ``A device that collects light is a telescope'' \\
\hline
\textit{assignment} & \textit{document} & \textit{student writes} & ``A document that a student writes is an \\
& & & assignment'' \\
& & \textit{student submits} & ``A document that a student submits is an \\
& & & assignment'' \\
& & \textit{teacher reads} & ``A document that a teacher reads is an  \\
& & & assignment'' \\
& & \textit{receives a grade} & ``A document that receives a grade is an \\
& & & assignment'' \\
\hline
\textit{ruin} & \textit{building} & \textit{archaeologist discovers} & ``A building that an archaeologist discovers \\
& & & is a ruin'' \\
& & \textit{dig excavates} & ``A building that a dig excavates is a ruin'' \\
& & \textit{archaeologist studies} & ``A building that an archaeologist studies is \\
& & & a ruin'' \\
& & \textit{collector restores} & ``A building that a collector restores is a ruin'' \\
& & \textit{jungle covers} & ``A building that the jungle covers is a ruin'' \\
& & \textit{excavation reveals} & ``A building that excavation reveals is a ruin'' \\
\hline
\end{tabular}
\caption{RELPRON dataset entries for the terms \textit{telescope}, \textit{assignment}, and \textit{ruin}, including their respective hypernyms, properties, and the verbalized templates derived from each (term, hypernym, property) triple.}
\label{table_relpron_ex}
\end{table*}

\begin{table}[h]
\centering
\scalebox{0.66}{\begin{tabular}{l|l}
\textbf{Sentence} & \textbf{Label} \\
\hline
\textit{A particular person didn't mean \underline{to do a particular thing}} & 1 \\
\textit{Someone didn't tell \underline{a particular person to do a particular thing}} & 0 \\
\textit{John wasn't upset that \underline{a particular thing happened}} & 1 \\
\textit{John didn't find that \underline{a particular thing happened}} & 0 \\
\textit{A particular person was thrilled \underline{to do a particular thing}} & 1 \\
\textit{A particular person yearned \underline{to have a particular thing}} & 0
\end{tabular}}
\caption{Examples of sentences (with subordinate clauses \underline{underlined}) and their corresponding labels from the MegaVeridicality 2.1 dataset. A label of 1 indicates that the subordinate event is portrayed as true, while a label of 0 indicates that is not.}
\label{table_factuality_ex}
\end{table}

\begin{table}[h]
\centering
\begin{tabular}{l|l}
\textbf{Feature} & \textbf{Value} \\
\hline
\textit{a-utensil} & 0.634 (19/30) \\
\textit{found-in-kitchens} & 0.600 (18/30) \\
\textit{used-with-forks} & 0.534 (16/30) \\
\textit{a-cutlery} & 0.500 (15/30) \\
\textit{is-dangerous} & 0.467 (14/30) \\
\textit{a-weapon} & 0.367 (11/30)
\end{tabular}
\caption{\cite{mcrae-etal-2005-semantic} feature norms for the concept \textit{knife} (feature values are obtained from experiment participants' judgments). For all other features $Q$, $F(knife)_Q = 0$.}
\label{table_mcrae_ex}
\end{table}

\newpage
\hspace{0mm}
\newpage

\begin{figure}[h]
\centering
\includegraphics[width=75mm, height=44mm]{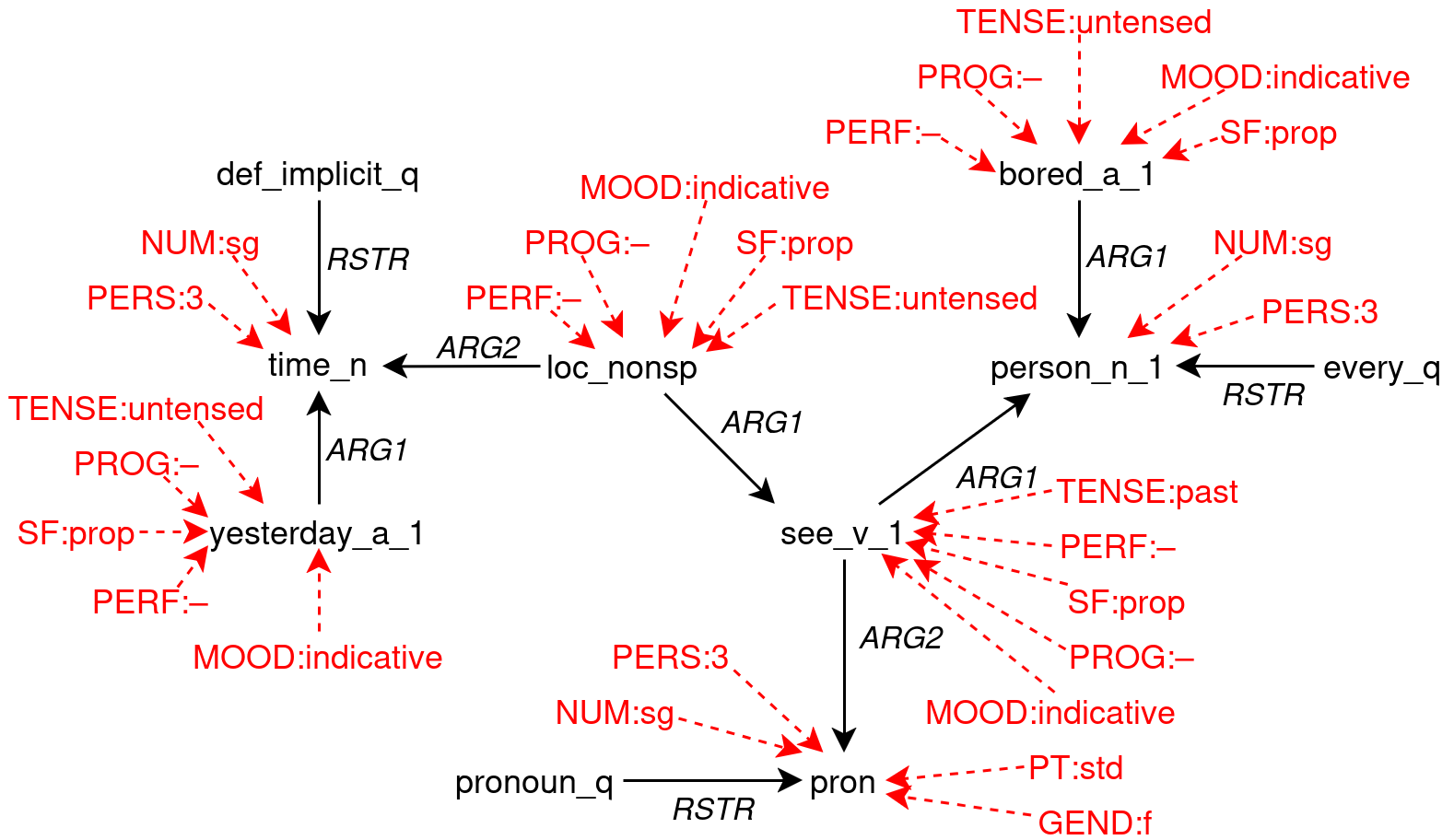}
\caption{DMRS graph from Figure \ref{fig_dmrs_ex}, with node features included. Features are highlighted in red; a dashed arrow $\phi\rightarrow x$ indicates that $\phi$ is a feature of the node $x$.}
\label{fig_dmrs_ex_feats}
\end{figure}

\newpage
\hspace{0mm}
\newpage
\hspace{0mm}
\newpage

\section{Encoder Architectural Details}
\label{app_archictecture}

As discussed in Section \ref{sec_model_sub_arch}, the outputs of the embedding layer and positional encoding network ($E(X,G)$ and $P(E(X,G),G)$, respectively) are summed together and passed to the encoder stack. 

\begin{figure}[t]
\centering
\includegraphics[width=75mm, height=28mm]{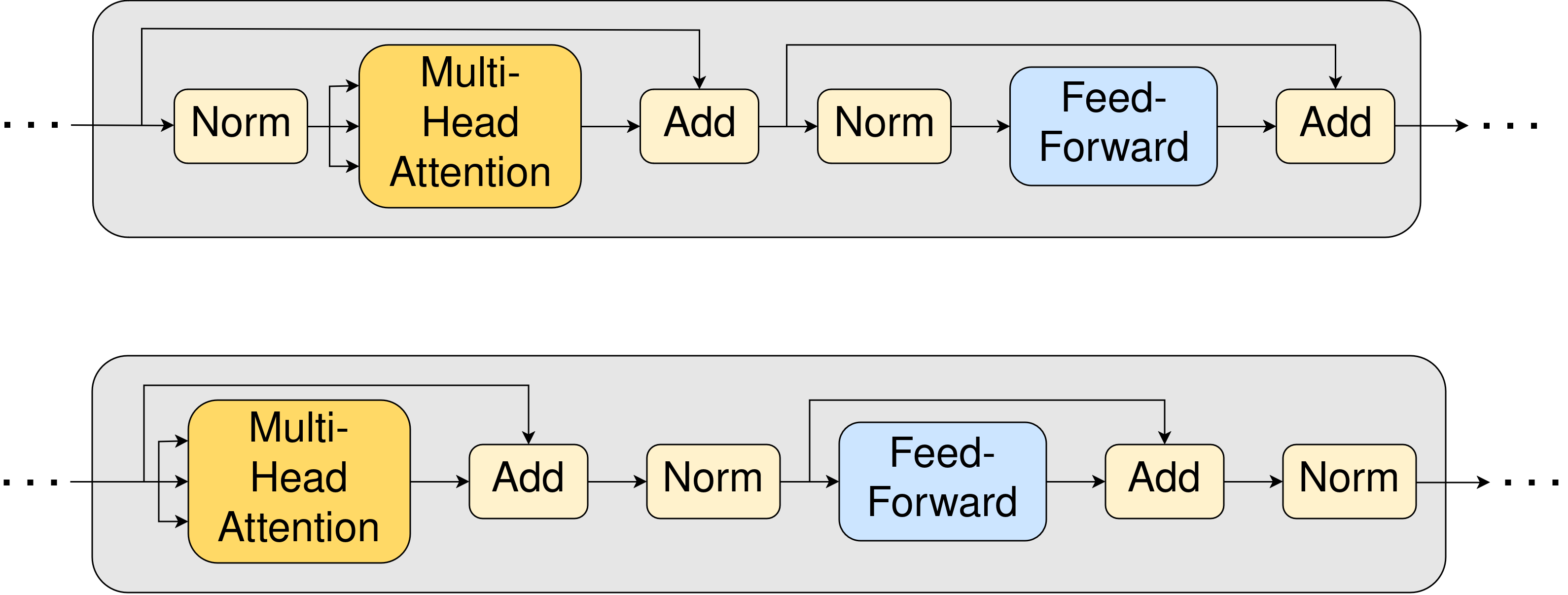}
\caption{Architecture of a GFoLDS (top) and BERT (bottom) encoder layer.}
\label{fig_gfolds_bert_enc}
\end{figure}

The encoder layers in the GFoLDS architecture are similar to those in BERT and \citet{vaswani2017}\footnote{Which is also the architecture employed in \cites{wu2021representing} original formulation of the graph transformer.}, but contain a few key differences\textemdash in particular with respect to the residual connections and interrelated layer normalization. As shown in Figure \ref{fig_gfolds_bert_enc}, in a BERT encoder layer, the layer input is directly passed to the multi-head attention module, immediately followed by a skip connection and layer normalization. The output of this first layer normalization is then passed to the feed-forward layer, which is again immediately followed by a skip connection and second layer normalization.

In a GFoLDS encoder layer, the input is first layer-normalized \textit{before} being passed to the multi-head attention module, which is followed by a skip connection. The next skip connection\textemdash in contrast to BERT\textemdash is copied \textit{before} layer normalization, which itself is followed by the feed-forward layer (which is identical to a BERT encoder feed-forward layer) and a second skip connection; this skip connection is \textit{not} followed by layer normalization. 

These architectural differences are motivated by the fact that\textemdash since the introduction of BERT\textemdash normalization \textit{outside} of the residual connection (i.e.\hspace{1mm}$\text{\textit{Norm}}(x+f(x))$) has been shown to be problematic. \citet{tuningplaybookgithub} instead recommend normalization \textit{inside} the residual (i.e.\hspace{1mm}$x+f(\text{\textit{Norm}}(x))$), which we implemented in GFoLDS.

\section{Data Preprocessing}
\label{app_preproc}

\begin{table*}[ht]
\centering
\begin{tabular}{|l|l|l|}
\hline
\textbf{Original Label} & \textbf{Interpretation/Role} & \textbf{Replacement} \\
\hline
$\text{\textit{ARG1}}$ & First-place argument & \textemdash \\
\hline
$\text{\textit{ARG2}}$ & Second-place argument  & \textemdash \\
\hline
$\text{\textit{ARG3}}$ & Third-place argument & \textemdash \\
\hline
$\text{\textit{ARG4}}$ & Fourth-place argument & \textemdash \\
\hline
$\text{\textit{MOD}}$ & Indicates a shared handle & \textemdash \\
& between two predicates & \\
\hline
$\text{\textit{RSTR}}$ & Restriction of a quantifier & \textemdash \\
\hline
$\text{\textit{ARG}}$ & Argument of the ``\textit{unknown}'' & $\text{\textit{MOD}}$ \\
& predicate & \\
\hline
$\text{\textit{L-INDEX}}$ & Left-hand conjunct of two & $\text{\textit{INDEX}}$ \\
& coordinated variables & \\
\hline
$\text{\textit{R-INDEX}}$ & Right-hand conjunct of two & $\text{\textit{INDEX}}$ \\
& coordinated variables & \\
\hline
$\text{\textit{L-HNDL}}$ & Left-hand conjunct of two & $\text{\textit{HNDL}}$ \\
& coordinated handles & \\
\hline
$\text{\textit{R-HNDL}}$ & Right-hand conjunct of two & $\text{\textit{HNDL}}$ \\
& coordinated handles & \\
\hline
\end{tabular}
\caption{DMRS edge labels (left), the role that they play in describing meaning (center), and the edge labels that they are replaced with (right) during preprocessing (if any: ``\hspace{1mm}\textemdash\hspace{1mm}'' indicates that a label is retained after preprocessing).}
\label{table_dmrs_edge_lbl}
\end{table*}

After removing CARGs and OOV items from the graph structures (see Section \ref{sec_model_sub_preproc}), there remain a few\textemdash relatively less significant\textemdash preprocessing steps that we took in order to transform DMRS representations into inputs for the GFoLDS model. 

\begin{figure}[t]
\centering
\includegraphics[width=75mm, height=19mm]{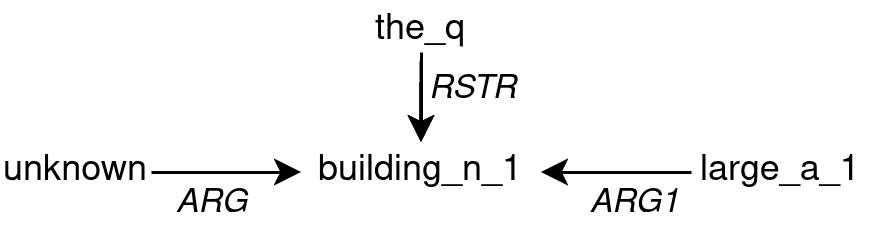}
\caption{DMRS representation of the noun phrase ``\textit{the large building}''.}
\label{fig_unk_pred_ex}
\end{figure}

For the sake of semantic well-formedness, the ACE/ERG parser attempts to represent all inputs as if they were entire sentences. For example, given the input ``\textit{the large building}'', the parser will parse the noun phrase, then insert the predicate \textit{unknown}\footnote{Not to be confused with out-of-vocabulary/``unknown'' items, which are represented in a different manner in DMRS.} and an edge $\text{\textit{ARG}}\colon\text{\textit{unknown}}\rightarrow\text{\textit{building\_n\_1}}$ (see Figure \ref{fig_unk_pred_ex}), which indicates that there is some unknown (presumably verbal) predicate for which the building plays an (again unknown) semantic role (indicated by $\text{\textit{ARG}}$). Such constructions are the \textit{only} context in which the $\text{\textit{ARG}}$ edge label appears in DMRS.

While this representational choice is sensible from the perspective of formal semantics, it is undesirable from the viewpoint of machine learning. The fact that $\text{\textit{ARG}}$ only links \textit{unknown} to other predicates (and is \textit{always} included when \textit{unknown} is in the graph) makes that predicate extremely predictable: if, during pretraining, the model is given a graph with a masked \textit{unknown} predicate, it needs only look for the $\text{\textit{ARG}}$-labeled edge to know that \textit{unknown} is the masked node. This means that the model does not need to learn any co-occurrence relations between \textit{unknown} and other nodes in the graph in order to learn to reliably predict the distribution of \textit{unknown}. 

Second, recall that each unique argument label is assigned unique forward and backward $d_{\text{\textit{SWA}}}\times d_{\text{\textit{SWA}}}$ edge projection layers in each SWA layer (see Section \ref{sec_model_sub_arch}). This is to say that each unique edge label corresponds to $2n(d_{\text{\textit{SWA}}})^2$ parameters in the GFoLDS model, where $n$ denotes the number of SWA layers. If, for example, $n=2$ and $d_{\text{\textit{SWA}}}=1024$, then each additional edge label adds 4,194,304 parameters to the model architecture. Given the highly specialized function of the $\text{\textit{ARG}}$ edge label, it seems rather unreasonable to allocate so many parameters to its representation. 

Therefore, during preprocessing, we converted each $\text{\textit{ARG}}$ label to the $\text{\textit{MOD}}$ label (see Table \ref{table_dmrs_edge_lbl}): another purely structural DMRS edge label that is used to indicate handle equality between predicates when other argument-label edges alone are insufficient to do so \citep{muszynska2020semantic}.

Additionally, we equivalence-classed argument labels involved in coordination structures (corresponding to logical conjunction and disjunction): for a predicate such as \textit{and\_c}, DMRS includes the argument labels $\text{\textit{L-HNDL}}\colon\text{\textit{and\_c}}\rightarrow X$ and $\text{\textit{R-HNDL}}\colon\text{\textit{and\_c}}\rightarrow Y$ ($\text{\textit{L-INDEX}}$ and $\text{\textit{R-INDEX}}$, respectively, when the conjuncts are variables rather than handles) denoting the left- and right-hand conjuncts $X$/$Y$ (respectively) of the coordinated structure. Logically, however, conjunction and disjunction are commutative operators: $\phi\land\psi=\psi\land\phi$ and $\phi\lor\psi=\psi\lor\phi$. Therefore, we replaced the edge labels $\text{\textit{L-HNDL}}$/$\text{\textit{R-HNDL}}$ and $\text{\textit{L-INDEX}}$/$\text{\textit{R-INDEX}}$ with $\text{\textit{HNDL}}$ and $\text{\textit{INDEX}}$ (respectively; see Table \ref{table_dmrs_edge_lbl}), ignoring the surface order of the conjuncts. This preprocessing step has the added benefit of reducing the overall size of the GFoLDS model by $4n(d_{\text{\textit{SWA}}})^2$ parameters, as discussed above.

Finally, we removed from the graph structures all instances of \textit{focus\_d} and \textit{parg\_d}: predicates with a purely discourse-pragmatic role, which indicate that the predicates that they modify are focus-topicalized or the subject of a passivized verb (respectively). 

\section{Pretraining Details}
\label{app_pretraining}

This section describes additional details and hyperparameters of the pretraining processes for GFoLDS (Section \ref{app_pretraining_sub_gfolds}) and the BERT comparison models used in Sections \ref{sec_experiments}-\ref{sec_hyp1} (Section \ref{app_pretraining_sub_bert}).

\subsection{GFoLDS}
\label{app_pretraining_sub_gfolds}

\paragraph{Pretraining Objective} During BERT's pretraining procedure, while 15\% of the input tokens are \textit{selected} for prediction, only 80\% of the selected tokens are masked: this is to account for the mismatch between the model's pretraining and fine-tuning distributions that arises from the fact that the [MASK] token only occurs during pretraining. However, for GFoLDS, the [MASK] token \textit{does} occur during fine-tuning as well, due to OOV items (as discussed in Section \ref{sec_model_sub_preproc}). Furthermore, \citet{wettig-etal-2023-mask} find that higher selection rates\textemdash and higher \textit{masking} rates\textemdash result in improved performance on downstream tasks, when compared to the selection and masking/replacement rates reported in \citet{devlin-etal-2019-bert}. For these reasons, we chose to mask 100\% of the selected tokens during pretraining, with the slightly higher selection probability of 20\%. 

The MNM prediction head that we used is identical to BERT's MLM prediction head (aside from the difference in vocabulary size): a
$d_{\text{\textit{model}}}\times d_{\text{\textit{model}}}$ linear layer, followed by GeLU activation, layer norm, and a $d_{\text{\textit{model}}}\times\text{22077}$ (the size of the vocabulary) linear layer.

\paragraph{Hyperparameters}

\begin{figure}[t]
\centering
\includegraphics[width=75mm, height=58mm]{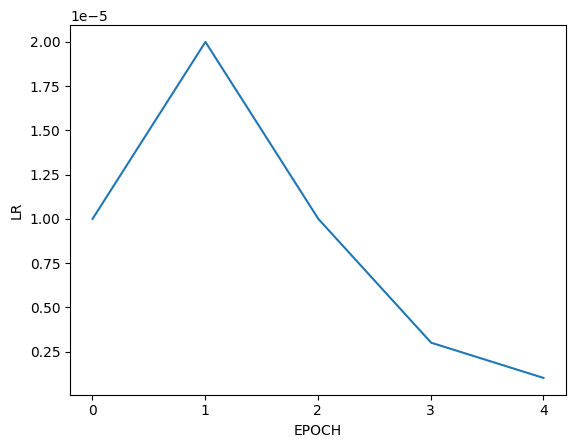}
\caption{GFoLDS pretraining learn rate schedule.}
\label{fig_lr_schedule_pt}
\end{figure}

We pretrained GFoLDS with a batch size of 16 for four epochs with the AdamW optimizer \cite{loshchilov2017decoupled} and a weight decay value of $10^{-5}$. We set an initial learn rate of $10^{-5}$, with a linearly interpolated learn rate between values of $2\times10^{-5}$ at the end of the first epoch, $10^{-5}$ at the end of the second, $3\times10^{-6}$ at the end of the third, and $10^{-6}$ at the end of the fourth (see Figure \ref{fig_lr_schedule_pt}). That is to say that the learn rate increased linearly from $10^{-5}$ to $2\times10^{-5}$ during the first epoch, decreased linearly from $2\times10^{-5}$ to $10^{-5}$ during the second epoch, and so on.

\begin{figure}[h]
\centering
\includegraphics[width=75mm, height=56mm]{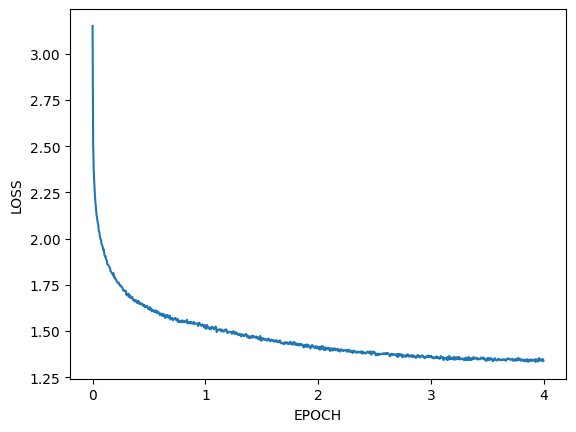}
\caption{GFoLDS pretraining cross-entropy loss.}
\label{fig_pt_loss}
\end{figure}

At the specified batch size of 16, the model trained at a rate of roughly 25 hours and 36 minutes per epoch on a single NVIDIA A100 GPU, for a total training time of 102 hours and 24 minutes. GFoLDS converged to a cross-entropy loss of $\sim$1.3331 at the end of the fourth epoch (see Figure \ref{fig_pt_loss}).

\subsection{BERT Comparison Models}
\label{app_pretraining_sub_bert}

\begin{figure*}[h]
\centering
\includegraphics[width=145mm, height=34mm]{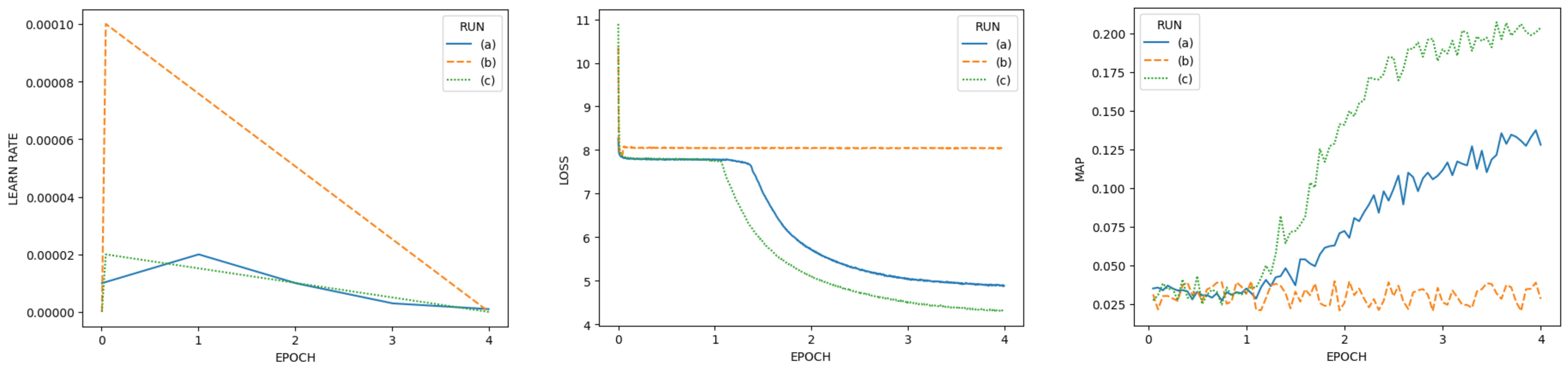}
\caption{Learning rates (left), cross-entropy loss values (center), and MAP scores on the RELPRON development split (right) across training steps for the $\text{BERT}_{\text{base}}$ pretraining trials (a)-(c).}
\label{fig_lr_loss_map_run1_3}
\end{figure*}

Given the differences in size and modality between the BERT and GFoLDS models, the best-performing set of pretraining hyperparameters for BERT on this dataset is not likely to be identical to those of GFoLDS. We therefore evaluated a variety of different hyperparameter configurations for BERT in order to yield the most equitable comparison with GFoLDS. Due to the relatively higher cost associated with training $\text{BERT}_{\text{large}}$ (over three times larger than $\text{BERT}_{\text{base}}$), we performed the majority of the trials with $\text{BERT}_{\text{base}}$, then transferred the best-performing configuration found during these experiments to $\text{BERT}_{\text{large}}$.

We first evaluated $\text{BERT}_{\text{base}}$ on three different configurations across which the learn rate schedule, weight decay, and masking rates varied: all three configurations employed the next sentence prediction (NSP) secondary pretraining task \cite{devlin-etal-2019-bert}. Due to hardware constraints, we were limited to using a batch size of 16 for all of the BERT pretraining trials. 

The first configuration (a) was identical to that which we employed for GFoLDS (see Appendix \ref{app_pretraining_sub_gfolds}): a weight decay value of $10^{-5}$; a linearly interpolated learn rate between values of $2\times10^{-5}$ at the end of the first epoch, $10^{-5}$ at the end of the second, $3\times10^{-6}$ at the end of the third, and $10^{-6}$ at the end of the fourth (with an initial learn rate of $10^{-5}$); and a token selection probability of 20\% with a masking probability of 100\%.

In the second configuration (b), we used a hyperparameter configuration that was identical to that of the original BERT models: a weight decay value of $10^{-2}$; linear learn rate warmup to $10^{-4}$ across the first 1\% of the training run (with linear decay thereafter); and a token selection probability of 15\% with a masking and replacement rates of 80\% and 10\%, respectively.

However, the original BERT models were pretrained with a batch size of 256, while trials (a) and (b) use a batch size of 16. Although we were not able to increase the batch size due to hardware constraints (as mentioned above), \citet{granziol2022learning} show that proportional (to batch size) learn rate scaling can be used to control for the effect of batch size on training loss. We therefore introduced a third trial (c): this configuration was identical to that of (b) with the exception of the peak learn rate value, which we scaled down to $2\times10^{-5}$ to account for the difference in batch size.

We then evaluated trials (a)-(c) with respect to learn rate and a validation task that does not require fine-tuning: the development split of the RELPRON \citep{rimell-etal-2016-relpron} dataset (see Section \ref{sec_hyp1_sub_setup_sub_relpron}). The results of these experiments are shown in Figure \ref{fig_lr_loss_map_run1_3}.

The model clearly failed to learn with the original BERT pretraining hyperparameters (b)\textemdash likely due to the mismatch in batch size discussed above\textemdash and finished training with a minimum cross-entropy loss of 7.8864 and a peak MAP score of 0.040 on the RELPRON development set. Of the two remaining configurations, (c) outperforms (a) both in terms of minimum cross-entropy (4.2988 vs.\hspace{1mm}4.8716) and peak MAP score (0.207 vs.\hspace{1mm}0.137).

\begin{figure}[h] 
\centering
\includegraphics[width=75mm, height=28mm]{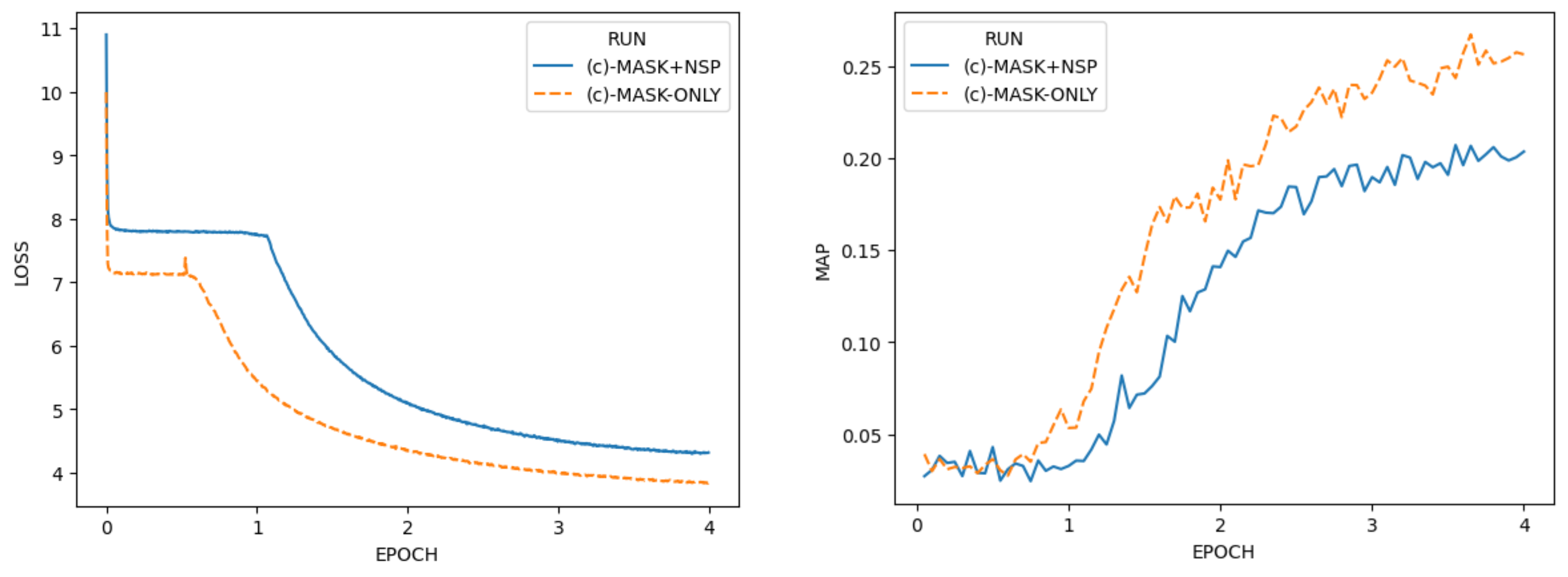}
\caption{Cross-entropy loss values (left), and MAP scores on the RELPRON development split (right) across training steps for the $\text{BERT}_{\text{base}}$ pretraining configuration (c) with and without the secondary NSP objective.}
\label{fig_loss_map_run3_4}
\end{figure}

\citet{liu2019roberta} suggest that pretraining with the secondary NSP objective does not improve (and in some cases may even hinder) model performance. We therefore conducted a fourth hyperparameter trial with $\text{BERT}_{\text{base}}$, using the same configuration as in (c) above, but excluding the NSP task. The results of this experiment are shown in Figure \ref{fig_loss_map_run3_4}.

The variant of configuration (c) without NSP outperforms the original trial in terms of cross-entropy loss (3.8339 vs.\hspace{1mm}4.2988, respectively; see Figure \ref{fig_loss_map_run3_4})\textemdash this is to be expected: the loss values reported for the variant with NSP are the sum of the NSP loss with the masked language modeling (MLM) loss. However, the non-NSP configuration also outperforms its NSP counterpart in terms of peak MAP score on the RELPRON development set: 0.267 vs.\hspace{1mm}0.207 (respectively). For this reason, we selected the non-NSP variant of the model pretrained with hyperparameter configuration (c) as the $\text{BERT}_{\text{base}}$ comparison model to be used in the experiments in this work.

\begin{figure*}[h]
\centering
\includegraphics[width=145mm, height=36mm]{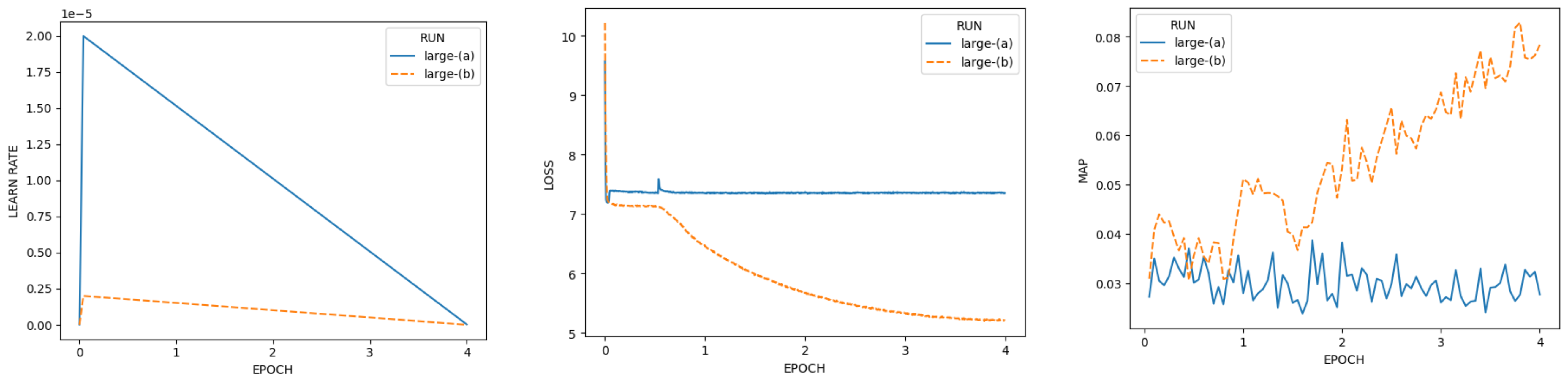}
\caption{Learning rates (left), cross-entropy loss values (center), and MAP scores on the RELPRON development split (right) across training steps for the $\text{BERT}_{\text{large}}$ pretraining trials (a)-(b).}
\label{fig_lr_loss_map_lrg}
\end{figure*}

We then pretrained $\text{BERT}_{\text{large}}$ with (non-NSP) hyperparameter configuration (c). As shown in Figure \ref{fig_lr_loss_map_lrg} (configuration large-(a)), this model failed to converge: it finished training with a minimum cross-entropy loss of 7.1856 and a peak MAP score of 0.039 on the RELPRON development set. As larger neural networks are more prone to overfitting \citep{caruana2000overfitting,salman2019overfitting}, we scaled the peak learn rate by a factor of $1/10$ in trial large-(b) to account for the difference in size between the $\text{BERT}_{\text{base}}$ and $\text{BERT}_{\text{large}}$ models. 

Trial large-(b) vastly outperformed large-(a) in terms of both minimum cross-entropy loss (5.1998 vs.\hspace{1mm}7.1856) and peak RELPRON development split MAP score (0.039 vs.\hspace{1mm}0.083), although it trails far behind the best-performing $\text{BERT}_{\text{base}}$ configuration by both metrics (3.8339 vs.\hspace{1mm}5.1998 cross-entropy; 0.267 vs.\hspace{1mm}0.083 RELPRON MAP). This substantial difference in performance between the base and large BERT variants is to be expected: it is well-known that larger neural networks require more training data in order to properly converge \citep[see e.g.][]{hoffmann2022training,muennighoff2024scaling}, and $\text{BERT}_{\text{large}}$ has over triple the amount of parameters (and double the number of encoder layers) as the base version of the model. Further compounding this issue is the fact that we pretrained these comparison models on 6.5 times less data (and for ten times fewer epochs) than was intended for the BERT models.

With a batch size of 16, $\text{BERT}_{\text{large}}$ trained at a rate of roughly 35 hours and 20 minutes per epoch on a single NVIDIA H100 GPU, for a total training time of 141 hours and 20 minutes. For comparision, GFoLDS and $\text{BERT}_{\text{base}}$\textemdash each trained on an NVIDIA A100\textemdash required 102 hours and 24 minutes, and 46 hours and 36 minutes, respectively (total training time). 

\section{SNLI Fine-Tuning Details}
\label{app_finetuning}

\begin{figure}[h]
\centering
\includegraphics[width=75mm, height=59mm]{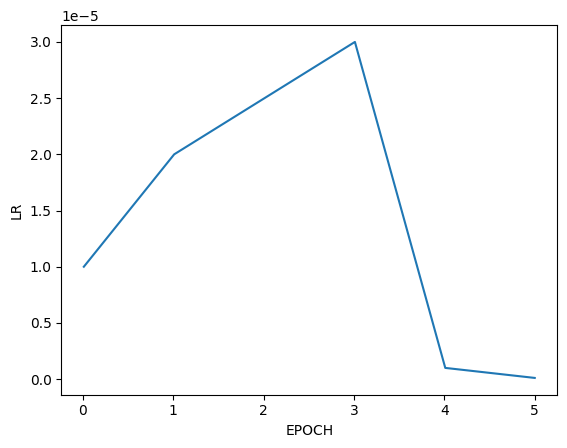}
\caption{SNLI fine-tuning learn rate schedule for GFoLDS and the $\text{BERT}_\text{base}$ models.}
\label{fig_lr_schedule_snli}
\end{figure}

We fine-tuned GFoLDS and $\text{BERT}_{\text{base}}$ for five epochs with a batch size of 16, a weight decay value of $10^{-5}$, an initial learn rate of $10^{-5}$, and a linearly-interpolated learn rate (updated at each batch) between values of $2\times10^{-5}$ at the end of the first epoch, $3\times10^{-5}$ at the end of the third, $10^{-6}$ at the end of the fourth, and $10^{-7}$ at the end of the fifth (see Figure \ref{fig_lr_schedule_snli}). We fine-tuned $\text{BERT}_{\text{large}}$ with identical hyperparameters, except all learn rates mentioned above were multiplied by 1/10 for this model ($\text{BERT}_{\text{large}}$ was unstable with the higher learn rate used for GFoLDS and $\text{BERT}_{\text{base}}$).

The fine-tuning hyperparameters for the BERT models were admittedly not selected using as rigorous of a search procedure as that employed during pretraining (see Appendix \ref{app_pretraining_sub_bert}). However, the accuracy that the (original) $\text{BERT}_\text{base}$ and $\text{BERT}_\text{large}$ models attained with the hyperparameter configurations described in the above paragraph (see Table \ref{table_downstream_res}) matched that reported for those models on the SNLI dataset in \citet{zhang2020semantics}. It is therefore reasonably safe to assume that this set of hyperparameters is (near-)optimal for the BERT models with respect to this data.

\begin{table*}[h]
\centering
\begin{tabular}{l|l}
\textbf{Type} & \textbf{Quantifiers} \\
\hline
\textit{SG} & another, either, neither, that, this, every, a(n), each \\
\textit{PL} & these, certain, most, those, all, such, both \\
\textit{BOTH} & some, the, any, enough, no, which \\
\end{tabular}
\caption{Quantifier types\textemdash along with a list of the quantifiers belonging to each type\textemdash used in the quantifier-agreement task.}
\label{table_quant_types}
\end{table*}

\section{MegaVeridicality V2.1 Task Setup}
\label{app_factuality}

We converted this dataset to a binary classification task by assigning a values of $1$, $0$, and $-1$ to the labels \textit{yes}, \textit{maybe}, and \textit{no} (respectively). We then assigned each example a value of 1 (i.e.\hspace{1mm}the subordinate event is portrayed as true) if its mean value was greater than zero, and 0 otherwise.

\section{Elementary Tasks}
\label{app_elementary}

\paragraph{POS-prediction} As discussed in Section \ref{sec_hyp1_sub_setup}, we evaluated the models on 200 sentences drawn from English Wikipedia. We first parsed each sentence using the ACE/ERG DMRS parser (see Section \ref{sec_model_sub_preproc}), which automatically labels the part-of-speech of each predicate in the DMRS representation of a sentence. We then randomly selected a single word to mask from each parsed sentence\textemdash subject to the condition that the selected word must be mapped to a single token by the BERT tokenizer (in order to facilitate the evaluation of the BERT comparison models)\textemdash and recorded to part-of-speech of the selected word. This resulted in a dataset consisting of 58 masked-quantifier, 28 masked-preposition, 33 masked-verb, 63 masked-noun, and 18 masked-adjective sentences. 

Determining the part-of-speech for each prediction of the GFoLDS model was trivial: DMRS predicates include part-of-speech tags, so we simply checked the tag of each predicted predicate. The BERT models, however, necessitated the use of the NLTK POS tagger\footnote{\href{https://www.nltk.org/api/nltk.tag.pos_tag.html}{https://www.nltk.org/api/nltk.tag.pos\_tag.html}}. For each sentence $s$, and each of the model's top-ten predicted tokens $w$ for $s$, we created a new sentence $s_w$ by replacing the masked word of $s$ with the prediction $w$, and ran the NLTK POS tagger over $s_w$ to obtain the tag for $w$. Note that, while the NTLK POS tagger is not perfect, it does achieve 95+\% accuracy on English-language data \citep{jacobsen2021optimal}, and therefore is sufficiently robust to yield an estimate of the model's performance. 

We chose to use (bounded) precision as the evaluation metric for this task because of the large amounts of positive examples for each class (especially nouns, verbs, and adjectives), which precluded the calculation of metrics that incorporate false negatives (e.g.\hspace{1mm}recall and F1). We recorded each model's mean precision across all 200 sentences as its final score for this task.

\paragraph{Quantifier-agreement} We first parsed each sentence with the ACE/ERG parser, which explicitly labels the number of each noun: this allowed the automatic extraction of the number of the noun in the restriction of a given quantifier. We then randomly selected a single quantifier from each sentence to mask, and recorded the number of the noun in the quantifier's restriction. 

We sorted all of the quantifiers into one of three categories/types (see Table \ref{table_quant_types}): \textit{singular} (can only restrict singular nouns), \textit{plural} (can only restrict plural nouns), and \textit{both} (can restrict either kind of noun). Note that the both-type quantifiers were used only for evaluation: the type of the masked quantifiers was recorded only as singular or plural\textemdash all nouns are either singular or plural, and the target type was determined by the number assigned by the ACE/ERG parser to the noun in the quantifier's restriction.

When computing precision for this task, all non-quantifier words in the models' top-$k$ predictions were treated as false positives, while both-type quantifiers in the top $k$ were treated as true positives, regardless of the target type (singular or plural). As in the POS-prediction task above, we recorded each model's
mean precision across all 179 sentences as its final score for this task.

\section{Scalability}
\label{app_scalability}

\subsection{Background and Notation}
\label{app_scalability_sub_background}

\cite{hoffmann2022training} investigate the relationship between model size (i.e.\hspace{1mm}number of parameters) and amount of pretraining data, and the final pretraining cross-entropy loss of LLMs. The optimal number of parameters and amount of data (number of tokens) for a fixed compute budget\textemdash expressed in terms of floating-point operations (FLOPs)\textemdash is given in Equation \ref{eq_hoffman_opt}, where $N_{opt}$ denotes the optimal number of model parameters, $D_{opt}$ the optimal number of training tokens, $C$ the compute budget, and $L(N,D)$ the model's final pretraining loss on $D$ tokens with $N$ parameters. 

\begin{equation}
\label{eq_hoffman_opt}
(N_{\text{\textit{opt}}},D_{\text{\textit{opt}}})=\underset{(N,D)\hspace*{0.5mm}:\hspace*{0.5mm}\text{\textit{FLOPs}}(N,D)\hspace*{0.25mm}=\hspace*{0.25mm}C} {{\text{\textit{argmin}}}} L(N,D)
\end{equation}

The authors fit a model of LLM final pretraining loss as a function of $N$ and $D$. This model is given in Equation \ref{eq_hoffman_est}, where $E$, $A$, $B$, $\alpha$, and $\beta$ are learned constants\footnote{$E=1.69$, $A=406.4$, $B=410.7$, $\alpha=0.34$, $\beta=0.28$}. 

\begin{equation}
\label{eq_hoffman_est}
L(N,D)\approx\hat{L}(N,D)=E+\frac{A}{N^\alpha}+\frac{B}{D^\beta}
\end{equation}

However, \cite{hoffmann2022training} only consider \textit{unique} training data\textemdash i.e.\hspace{1mm} pretraining a language model for a single epoch. \cite{muennighoff2024scaling} extend \cites{hoffmann2022training} experiments to the case of \textit{repeated} training data: pretraining the model for multiple epochs on the same dataset. The authors fit an analogous loss-prediction function to that given in Equation \ref{eq_hoffman_est} (Equation \ref{eq_muennighoff}a), $\hat{L}^{(r)}(N,U_D)$, which estimates the final loss for a language model with $N$ parameters trained on $U_D$ tokens for $r$ repetitions (i.e.\hspace{1mm}epochs).

\begin{subequations}
\label{eq_muennighoff}
\begin{align}
&L^{(r)}(N,U_D)\approx\hat{L}^{(r)}(N,U_D)=E+\frac{A}{\hat{N}^\alpha}+\frac{B}{\hat{D}^\beta} \\
&\hat{D}=U_D+U_DR^{*}_D(1-e^{-r/R^{*}_D}) \\
&\hat{N}=U_N+U_NR^{*}_N(1-e^{-R_N/R^{*}_N}) \\
&U_N=\text{\textit{min}}\{N,N_{\text{\textit{opt}}}(U_D)\} \\
&R_N=\frac{N}{U_N}-1
\end{align}
\end{subequations}

Where $U_D$ denotes the number of \textit{unique} tokens (i.e.\hspace{1mm}the amount of tokens in a single epoch), $N_{\text{\textit{opt}}}(U_D)$ is estimated as in \cite{hoffmann2022training} (see Equations \ref{eq_hoffman_opt} and \ref{eq_hoffman_est}), and $E$, $A$, $B$, $\alpha$, $\beta$, $R^{*}_D$, and $R^{*}_N$ are learned constants\footnote{$E=1.88$, $A=523.22$, $B=1480.30$, $\alpha=0.35$, $\beta=0.35$, $R^{*}_D=15.39$, $R^{*}_N=5.31$}.

\subsection{Proof}
\label{app_scalability_sub_proof}

The term $D$ in Equation \ref{eq_muennighoff} denotes the number of pretraining \textit{tokens}. This metric is likely not a perfect predictor for GFoLDS: recall from Section \ref{sec_model_sub_arch} that the model contains separate projection layers for each DMRS edge label type. The model therefore receives training signal not only from the node labels (i.e.\hspace{1mm}tokens), but also from the \textit{edge} labels. For example, an input graph with six tokens and five edges will update fewer model parameters than a graph with six tokens and ten edges. 

It is beyond the scope of this work to establish exact scaling laws for GFoLDS and determine the graph-based analogue to the term $U_D$ in Equation \ref{eq_muennighoff}. However, it is clear that such a $U_D$ value scales (more or less) linearly with the number of graphs in the pretraining dataset: as with a textual model, $U_D$ can be expressed as the sum of the amounts of data (regardless of how it is quantified) contributed by each individual input graph (respectively, sequence) in the dataset.

It is therefore reasonable to assume that $U_{D'}\approx U_D/2$, where $U_D$ denotes the dataset for the 100\% run, and $U_{D'}$ that of the 50\% run (see Section \ref{sec_scalability}). We may express the relationship between the 50\% run loss and the 100\% loss in the notation introduced in Appendix \ref{app_scalability_sub_background} (while leaving the exact definition of $U_D$ to future work): $L^{(4)}(N,U_D)\approx L^{(4)}(N,U_D/2)$. Recall from Equation \ref{eq_muennighoff}a that the term $E$ is a constant, and so can be ignored. We are then left with the following (approximate) equality in Equation \ref{eq_scale_eq}a.

\begin{subequations}
\label{eq_scale_eq}
\begin{align}
&\frac{A}{\hat{N}^{\alpha}_1}+\frac{B}{\hat{D}^{\beta}_1}\approx\frac{A}{\hat{N}^{\alpha}_2}+\frac{B}{\hat{D}^{\beta}_2} \\
&\hat{D}_1=U_D+U_DR^{*}_D(1-e^{-4/R^{*}_D}) \\
&\hat{D}_2\approx(U_D/2)+(U_D/2)R^{*}_D(1-e^{-4/R^{*}_D}) \\
&\hat{N}_1=U^{(1)}_N+U^{(1)}_NR^{*}_N(1-e^{-R^{(1)}_N/R^{*}_N}) \\
&\hat{N}_2=U^{(2)}_N+U^{(2)}_NR^{*}_N(1-e^{-R^{(2)}_N/R^{*}_N}) \\
&U^{(1)}_N=\text{\textit{min}}\{N,N_{\text{\textit{opt}}}(U_D)\} \\
&U^{(2)}_N\approx\text{\textit{min}}\{N,N_{\text{\textit{opt}}}(U_D/2)\} \\
&R^{(k)}_N=\frac{N}{U^{(k)}_N}-1
\end{align}
\end{subequations}

Where the left-hand expression in Equation \ref{eq_scale_eq}a corresponds to the 100\% run ($\hat{L}^{(4)}(N,U_D)$), and the right-hand expression corresponds to the 50\% run ($\hat{L}^{(4)}(N,U_D/2)$). Note that $\hat{D}_2\approx\hat{D}_1/2$ (Equation \ref{eq_d}).

\begin{equation}
\label{eq_d}
\begin{split}
   \hat{D}_2&\approx\frac{U_D}{2}+\frac{U_D}{2}R^{*}_D(1-e^{-4/R^{*}_D}) \\
   &=\frac{U_D+U_DR^{*}_D(1-e^{-4/R^{*}_D})}{2}=\frac{\hat{D}_1}{2}
\end{split}
\end{equation}

Given the vast architectural and modal differences between GFoLDS and textual transformer models, we do not assume that the coefficients $E$, $A$, $B$, $\alpha$, $\beta$, $R^{*}_D$, and $R^{*}_N$ fitted in \cite{muennighoff2024scaling} are identical for GFoLDS. Moreover, as stated above, it is beyond the scope of this work to establish exact scaling laws for this model. However, from the fact that final loss decreases as pretraining data increases from the 3.125\% to the 50\% runs (see Section \ref{sec_scalability_sub_results}), we know that the coefficients $B$ and $\beta$ in \ref{eq_scale_eq} must be positive: this\textemdash in conjunction with the (approximate) equality in Equation \ref{eq_d}\textemdash means that it cannot be the case that $B/\hat{D}^{\beta}_1>B/\hat{D}^{\beta}_2$.

Assume that $B/\hat{D}^{\beta}_1\ll B/\hat{D}^{\beta}_2$: given the equality in Equation \ref{eq_scale_eq}a, it must then be the case that $A/\hat{N}^{\alpha}_1-A/\hat{N}^{\alpha}_2\approx B/\hat{D}^{\beta}_2-B/\hat{D}^{\beta}_1$, and therefore that $A/\hat{N}^{\alpha}_2\ll A/\hat{N}^{\alpha}_1$. 

Assume further that the model is underparameterized at 100\% of the data (i.e.\hspace{1mm}that $N<N_\text{\textit{opt}}(U_D)$): then $U^{(1)}_N=\text{\textit{min}}\{N,N_\text{\textit{opt}}(U_D)\}=N$ and $R^{(1)}_N=(N/U^{(1)}_N)-1=0$, which implies that $\hat{N}_1=U^{(1)}_N+U^{(1)}_N(1-1)=U^{(1)}_N=N$ (see Equation \ref{eq_scale_eq}d). As the terms $A$ and $\alpha$ are constants in Equation \ref{eq_muennighoff}a (and therefore in Equation \ref{eq_scale_eq}a), we may reduce to the inequality expressed in Equation \ref{eq_n_under} (where the last logical equivalence is by definition of $R^{(2)}_N$; see Equation \ref{eq_scale_eq}h).

\begin{equation}
\label{eq_n_under}
\begin{split}
   &\hat{N}^{-\alpha}_1\gg\hat{N}^{-\alpha}_2\leftrightarrow\hat{N}_1\ll\hat{N}_2 \\ 
   &\leftrightarrow N\ll U^{(2)}_N+U^{(2)}_NR^{*}_N(1-e^{-R^{(2)}_N/R^{*}_N}) \\
   &\leftrightarrow\frac{N-U^{(2)}_N}{U^{(2)}_N}\ll R^{*}_N(1-e^{-R^{(2)}_N/R^{*}_N}) \\
   &\leftrightarrow\frac{N}{U^{(2)}_N}-1\ll R^{*}_N(1-e^{-R^{(2)}_N/R^{*}_N}) \\
   &\leftrightarrow R^{(2)}_N/R^{*}_N\ll1-e^{-R^{(2)}_N/R^{*}_N} \\
\end{split}
\end{equation}

But this is impossible: by Bernoulli's inequality, $1+x\leq e^x$ for all $x$. This implies that $x\leq e^x-1$, which in turn implies $-x\leq 1-e^x$, which then implies $x>1-e^{-x}$. As it therefore cannot be the case that $R^{(2)}_N/R^{*}_N<1-e^{-R^{(2)}_N/R^{*}_N}$, we know that the model cannot be underparameterized if $B/\hat{D}^{\beta}_1\ll B/\hat{D}^{\beta}_2$.

Now assume that the model is either over- or well-parameterized at 100\% of the data (i.e.\hspace{1mm}that $N\geq N_\text{\textit{opt}}(U_D)$): then $U^{(1)}_N=\text{\textit{min}}\{N,N_\text{\textit{opt}}(U_D)\}=N_\text{\textit{opt}}(U_D)$. It is reasonable to assume that $N_\text{\textit{opt}}(U_D/2)<N_\text{\textit{opt}}(U_D)$ (i.e.\hspace{1mm}that the model's optimal number of parameters scales monotonically with the amount of pretraining data), which implies that $U^{(2)}_N=\text{\textit{min}}\{N,N_\text{\textit{opt}}(U_D/2)\}=N_\text{\textit{opt}}(U_D/2)<N_\text{\textit{opt}}(U_D)=U^{(1)}_N$. With a fixed number of parameters $N$, $\hat{N}$ (Equations \ref{eq_muennighoff}c, \ref{eq_scale_eq}d-e) is a monotonic function of $U_N$ \citep[by construction;][]{muennighoff2024scaling}, so we have $U^{(1)}_N>U^{(2)}_N\rightarrow\hat{N}_1>\hat{N}_2$.

But this implies that $A/\hat{N}^{\alpha}_1<A/\hat{N}^{\alpha}_2$. This\textemdash along with the assumption that $B/\hat{D}^{\beta}_1\ll B/\hat{D}^{\beta}_2$\textemdash contradicts the observed result that $L^{(4)}(N,U_D)\approx L^{(4)}(N,U_D/2)$, and so we know that the model cannot be well- or over-parameterized if $B/\hat{D}^{\beta}_1\ll B/\hat{D}^{\beta}_2$.

Given that it cannot be the case that the model of the 100\% run is under-, well-, or over-parameterized if $B/\hat{D}^{\beta}_1\ll B/\hat{D}^{\beta}_2$ (and that it must be one of the three), it therefore cannot be the case that $B/\hat{D}^{\beta}_1\ll B/\hat{D}^{\beta}_2$. As argued above, the coefficients $B$ and $\beta$ must be positive, so it also cannot be the case that $B/\hat{D}^{\beta}_1\gg B/\hat{D}^{\beta}_2$: we must then conclude that $B/\hat{D}^{\beta}_1\approx B/\hat{D}^{\beta}_2$. As $\hat{D}_1\approx2\hat{D}_2$ (see Equation \ref{eq_d}), we know that $B/\hat{D}^{\beta}_1\approx (B/\hat{D}^{\beta}_2)/2^\beta$, (and that $2^\beta>1$, as $\beta$ must be positive). It must therefore be the case that $B/\hat{D}^{\beta}_1\approx0$.

Given that $B/\hat{D}^{\beta}_1\approx B/\hat{D}^{\beta}_2$, the equality in Equation \ref{eq_scale_eq}a implies that $A/\hat{N}^{\alpha}_1\approx A/\hat{N}^{\alpha}_2$, which in turn implies that $\hat{N}_1\approx\hat{N}_2$. If GFoLDS were over- or well-parameterized at 100\% of the data, then the monotonicity of $\hat{N}$ would imply that $\hat{N}_1>\hat{N}_2$ (as discussed above), contradicting the conclusion that $\hat{N}_1\approx\hat{N}_2$.

On the other hand, replacing the inequalities in Equation \ref{eq_n_under} with (approximate) equalities yields $R^{(2)}_N/R^{*}_N\approx1-e^{-R^{(2)}_N/R^{*}_N}$: note that the only value of $x$ for which $x=1-e^{-x}$ is 0. If GFoLDS is underparameterized at the 50\% run, then $U^{(2)}_N=\text{\textit{min}}\{N,N_\text{\textit{opt}}(U_D/2)\}=N$, which implies that $R^{(2)}_N=0$, which in turn implies that $R^{(2)}_N/R^{*}_N=0$. Given that $N_\text{\textit{opt}}(U_D)$ is (likely) greater than $N_\text{\textit{opt}}(U_D/2)$, it is also the case that $R^{(1)}_N/R^{*}_N=0$ and $U^{(1)}_N=\text{\textit{min}}\{N,N_\text{\textit{opt}}(U_D)\}=N=U^{(2)}_N$: this implies that $A/\hat{N}^{\alpha}_1=A/\hat{N}^{\alpha}_2$ (see Equations \ref{eq_scale_eq}d and \ref{eq_scale_eq}e), which is congruent with the conclusion that $\hat{N}_1\approx\hat{N}_2$.

\end{document}